\documentclass{article}

\usepackage[final]{corl_2020} % Uncomment for the camera-ready ``final'' version.
\usepackage{amsmath,amsfonts,bm}

\def\eqref#1{equation~\ref{#1}}
\def\1{\bm{1}}

\DeclareMathAlphabet{\mathsfit}{\encodingdefault}{\sfdefault}{m}{sl}
\SetMathAlphabet{\mathsfit}{bold}{\encodingdefault}{\sfdefault}{bx}{n}

\usepackage{booktabs}       % professional-quality tables
\usepackage{microtype}      % microtypography
\usepackage{graphicx}
\usepackage{algorithm}
\usepackage{algorithmic}
\usepackage{multirow}
\usepackage{wrapfig}
\usepackage[font=small]{caption}

\usepackage{bibunits}
\defaultbibliography{main}
\defaultbibliographystyle{corlabbrvnat}

\newcommand\aes{GAIL+AES}
\newcommand\videolink{https://youtu.be/46rSpBY5p4E}

\graphicspath{{figures/}}

\title{Task-Relevant Adversarial Imitation Learning}

\author{Konrad \.Zo\l{}na$^{1,2,}$\thanks{Equal contribution. Work done at DeepMind. Corresponding authors: \texttt{\{kondiz,reedscot\}@google.com}.}
\And
Scott Reed$^{2,\ast}$
\And
Alexander Novikov$^{2}$
\And
Sergio G\'{o}mez Colmenarejo$^{2}$
\And
David Budden$^{2}$
\And
Serkan Cabi$^{2}$
\And
Misha Denil$^{2}$
\And
Nando de Freitas$^{2}$
\And
Ziyu Wang$^{2}$
\AND
$^{1}${\normalfont Jagiellonian University} \mbox{ } \mbox{ } \mbox{ } \mbox{ } \mbox{ }
$^{2}${\normalfont DeepMind}
}

\begin{document}
\maketitle
\vspace{-0.3cm}
\begin{abstract}
We show that a critical vulnerability in adversarial imitation is the tendency of discriminator networks to learn spurious associations between visual features and expert labels. When the discriminator focuses on task-irrelevant features, it does not provide an informative reward signal, leading to poor task performance. We analyze this problem in detail and propose a solution that outperforms standard Generative Adversarial Imitation Learning (GAIL). Our proposed method, Task-Relevant Adversarial Imitation Learning (TRAIL), uses constrained discriminator optimization to learn informative rewards. In comprehensive experiments, we show that TRAIL can solve challenging robotic manipulation tasks from pixels by imitating human operators without access to any task rewards, and clearly outperforms comparable baseline imitation agents, including those trained via behaviour cloning and conventional GAIL.
\end{abstract}

\keywords{Adversarial Imitation, Robot Manipulation} 

\begin{bibunit}

\section{Introduction}
Generative Adversarial Networks (GANs) have been remarkably successful in image generation~\citep{goodfellow2014generative,brock2018large}, and have inspired similar approaches to imitate behaviour.
In Generative Adversarial Imitation Learning (GAIL)~\citep{ho2016generative}, a discriminator network is trained to distinguish
agent and expert behaviour through its observations, and is then used as a reward function.
GAIL agents can overcome the exploration challenge by taking advantage of expert demonstrations, while also potentially achieving high asymptotic performance.
Despite these attractive properties, GAIL has not had the same impact as GANs. In particular, robust adversarial imitation from pixels for control applications such as robotic manipulation remains an open challenge.

\begin{wrapfigure}{r}{0.465\textwidth}
\vspace{-1.0cm}
  \centering
  \includegraphics[width=0.99\linewidth]{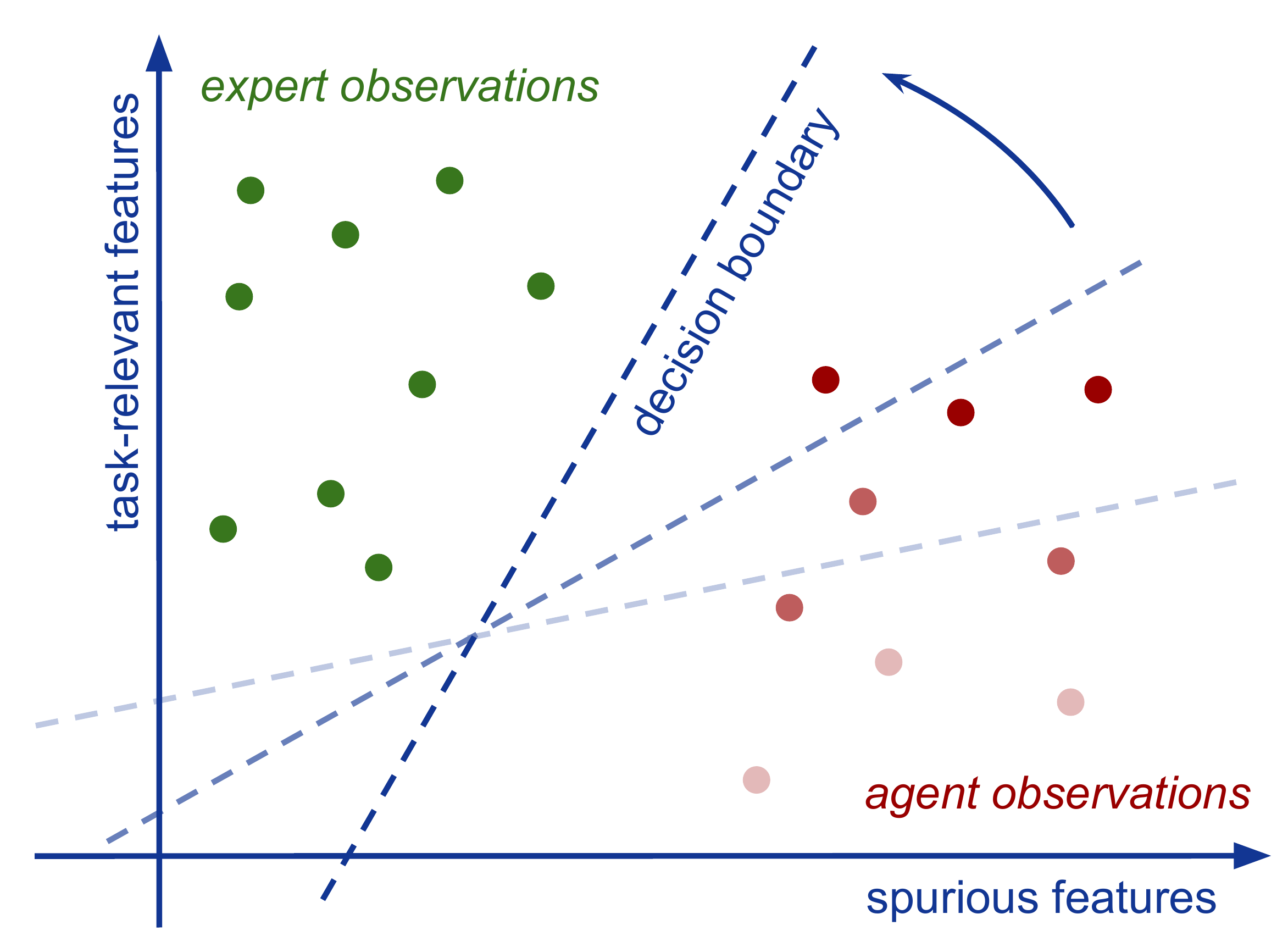}
  \caption{The decision boundary generated by a GAIL discriminator based on both task-relevant and spurious features. As the agent improves (more intense red dots), it produces observations closer to the expert w.r.t. task-relevant features. As a result, the discriminator decision boundary must increasingly rely on spurious features.}
  \vspace{-1.0cm}
  \label{fig:boundary_adaptation}
\end{wrapfigure}
One important reason for this challenge is causal confusion in the GAIL discriminator.
Given limited expert data and high-dimensional observations, the discriminator tends to exploit spurious associations between task-irrelevant features and expert labels.

To illustrate the problem, consider a simplified case where a discriminator has two feature sets. One is task-relevant, genuinely indicative of performance. The other features will be referred to as spurious, since they result in spurious associations in the discriminator.
See Figure~\ref{fig:boundary_adaptation} for a visualization.

At the start of training, the discriminator can use either task-relevant or spurious features to accurately distinguish the agent from expert observations.
However, as the agent improves, the task-relevant features become less predictive, which forces the discriminator to rely more on spurious features. 
This causes the rewards to become uninformative, and the agent performance therefore degrades.
The problem is especially challenging when some of the spurious features are beyond agent control (e.g. initial conditions or overall brightness).
Note that GAN generators fully control the discriminator input and hence, the problem of spurious associations may be more severe in the motor control setting as compared to image GAN training.

Even in low-dimensional settings, care must be taken to account for spurious associations.
For example, when using GAIL to learn simulated humanoid controllers from motion capture~\citep{merel2017learning}, there is a distributional shift that the discriminator could mistakenly associate with expert behavior.
Learning from pixels and adding props further compounds the problem, making robotic manipulation a particularly challenging domain for adversarial imitation.

Can the problem of spurious association in discriminator networks be overcome using generic regularization or data augmentation alone?
Research in the field of causality~\citep{simon1954spurious,pearl2010introduction} suggests not, in the general case.
If a model is not identifiable because of confounding, additional assumptions are needed in order to estimate causal effects \citep{pearl2009causality,peters2017elements}.  

We propose to introduce such assumptions in a flexible way.
Specifically, we regularize the discriminator to be unable to distinguish between \emph{constraining sets}, which consists of expert or agent observations, such that its elements can \emph{only} be identified as belonging to one or the other using spurious features.
This data-driven approach discourages the discriminator from forming spurious associations and, importantly, does not require  enumerating all possible spurious features.

We do not attempt to automate the choice of constraining sets for all possible tasks.
Rather, we show that our proposed approach can drastically improve agent performance in several challenging simulated robotic manipulation tasks.
Other problem settings might require constructing different constraint sets, but our proposed approach would still apply.

\begin{figure}[t]
  \vspace{-0.3cm}
  \centering
  \includegraphics[width=\linewidth]{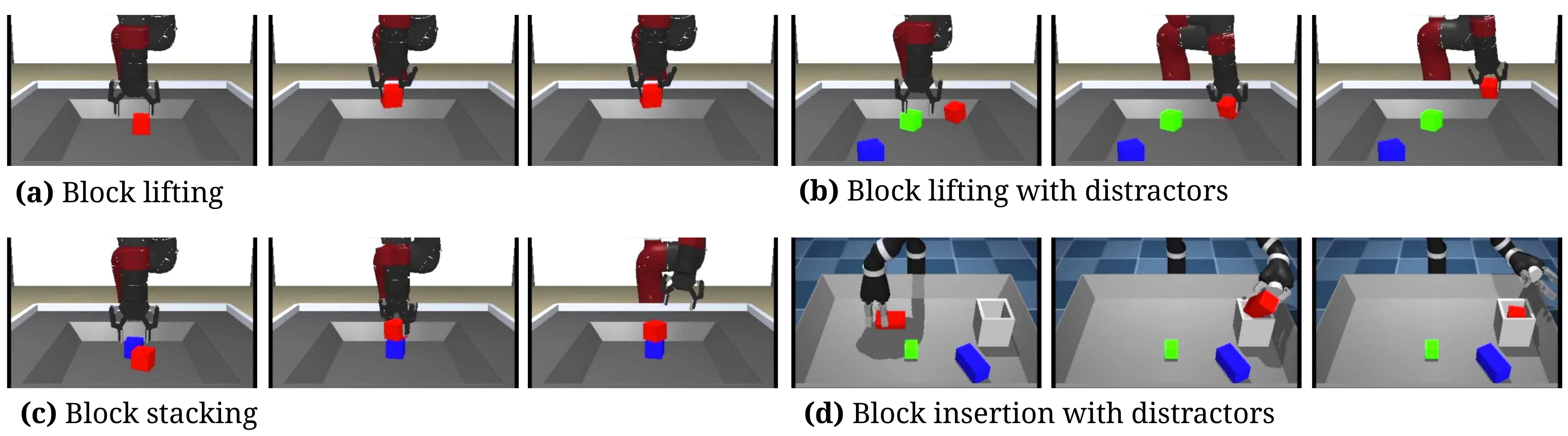}
  \caption{TRAIL agents solving a variety of manipulation tasks, including with distractor objects (b,d). A comparable GAIL agent can solve lifting (a) but fails when distractor objects are added (b-d), as additional objects trigger the formation of spurious associations in the discriminator. A video showing TRAIL and GAIL agents performing these tasks can be watched at \url{\videolink}.
  \label{fig:trail_tasks}}
  \vspace{-0.2cm}
\end{figure}

This paper makes the following contributions:\vspace{-0.35pc}
\begin{enumerate}
\item Highlights a fundamental problem in adversarial imitation, showing that GAIL discriminators do in practice exploit spurious associations, decreasing task performance.\vspace{-0.1pc}
\item Introduces \emph{Task-Relevant Adversarial Imitation Learning} (TRAIL), which effectively constrains the discriminator to focus on task-relevant patterns.\vspace{-0.1pc}
\item Improve performance on several vision-based robotic manipulation tasks (see Figure~\ref{fig:trail_tasks}).\vspace{-0.1pc}
\end{enumerate}
\vspace{-0.3pc}

\section{Related work}
\label{sec:related}
The use of demonstrations to help agent training has been studied extensively in robotics~\citep{bakker1996robot,kawato1994teaching,miyamoto1996kendama} with approaches ranging from Q-learning~\citep{schaal1997learning} to behavioral cloning (BC)~\citep{pomerleau1989alvinn}.
\textbf{Behavioral Cloning}\hspace{0.2cm}
BC is effective in solving many control problems \citep{pomerleau1989alvinn, finn2017one, duan2017one, rahmatizadeh2018vision}. It has also been successfully applied to initialize reinforcement learning (RL) training \citep{rajeswaran2017learning}. However, BC is vulnerable to cascading errors~\citep{ross2011reduction}.
This often necessitates a large number of demonstrations for satisfactory performance. 
Furthermore, BC agents typically cannot exceed demonstrator performance.

\textbf{Inverse RL}\hspace{0.2cm}
Inverse reinforcement learning (IRL) as a way of learning reward functions from demonstrations~\citep{ziebart2008maximum, ng2000algorithms, abbeel2004apprenticeship}.
RL can then be used to optimize that learned reward.
Recently, \citet{finn2016guided} approached continuous robotic control problems with success by applying Maximum Entropy IRL algorithms which are very closely related to GAIL \citep{finn2016connection} and have similar drawbacks.

\textbf{Learning from Demonstrations}\hspace{0.2cm}
\citet{hester2018deep} developed Deep Q-Learning from Demonstration (DQfD), in which expert trajectories are added to experience replay and jointly used to train agents along with their own experiences.
This was later extended to better handle sparse-reward problems~\citep{vecerik2017learning,pohlen2018observe}.
Despite their efficiency, these methods still require access to rewards for learning.

\textbf{Confounders and Causality}\hspace{0.2cm}
The study of spurious associations and causality is vast, so here we refer to the most relevant recent work.
In imitation learning, interventional approaches have been recently considered \cite{de2019causal}. In non-interventional settings, instrumental variables and proxy variables are 
two popular approaches for adding extra information into models to increase their identifiability. Instrumental variables have played an important role in offline reinforcement learning \cite{bradtke1996linear,lagoudakis2003least} and proxy variables have gained recent momentum in machine learning \cite{louizos2017causal,lu2018deconfounding,tennenholtz2019offpolicy}. In this paper, we propose an alternative data-driven solution, which is both simple and effective.    

\textbf{GAIL}\hspace{0.2cm}
Following the success of GANs in image generation, GAIL~\citep{ho2016generative} applies adversarial learning to the problem of imitation. Although many variants are introduced in the literature~\citep{li2017infogail,fu2017learning, merel2017learning,zhu2018reinforcement,baram2017end,feryal2019}, 
making GAIL work for high-dimensional input spaces, particularly raw pixels, remains a challenge.

The problem of forming spurious associations may resemble the problem of the discriminator overfitting which has been addressed before~\citep{peng2018variational, reed2018visual, blonde2018sample}.
Unstructured regularization, however, cannot stop the discriminator from fitting to visual features that are systematically different between the agent and the demonstrator.
Often it may be easier for a discriminator to exploit spurious features compared to task-relevant features, in which case strong regularization would intensify the problem.

\citet{stadie2017third} extend GAIL to the setting of third person imitation, in which the demonstrator and agent observations come from different views.
To prevent the discriminator from exploiting viewpoint, gradient flipping from an auxiliary classifier is used to learn domain-invariant features.
However, in our robotic manipulation setting, we found that the proposed method of extracting domain-invariant discriminator features did not adequately prevent the formation of spurious associations.
Unlike in~\citet{stadie2017third}, our goal is not third-person or cross-domain imitation, but rather a visual imitation method that is robust to spurious associations that could form even within a \emph{single} domain.

Several recent works have focused on improving the sample efficiency of GAIL~\citep{blonde2018sample, sasaki2018sample}.
Common to these approaches and to this work, is the use of off-policy actor critic agents and experience replay, to improve the utilization of available experience.

\section{RL and Adversarial Imitation Background}
\label{sec:background}

Following the notation of \citet{sutton2018reinforcement}, a Markov Decision Process (MDP) is a tuple $(\mathcal{S}, \mathcal{A}, R, P, \gamma)$ with states $\mathcal{S}$, actions $\mathcal{A}$, reward function $R(s,a)$, transition distribution $P(s' | s, a)$, and discount $\gamma$.
An agent in state $s \in S$ takes action $a \in A$ according to its policy $\pi$ and moves to state $s' \in \mathcal{S}$ according to the transition distribution.
The goal is to find a policy that maximizes the expected sum of discounted rewards, represented by the action value function $Q^{\pi}(s,a) = \mathbb{E}^{\pi}[\sum_{t=0}^{\infty}\gamma^t R(s_t, a_t)]$, where $\mathbb{E}^{\pi}$ is an expectation over trajectories starting from $s_0 = s$ and taking action $a_0 = a$ and thereafter running the policy $\pi$.

To apply RL, it is essential to have access to the reward function which is often hard to design and evaluate \citep{singh2019end}. In addition, sparse rewards can cause exploration difficulties that pose great challenges to RL algorithms. We therefore look to adversarial imitation learning to derive a reward function from expert demonstrations.

In adversarial imitation, a reward function is learned by training a discriminator network $D$ to distinguish between agent and expert states, or optionally state-action pairs.
In the state only case, the discriminator objective is:
\begin{align}
    \underset{\psi}{ \max} \texttt{ } \mathbb{E}_{s \sim \pi_E}[\log D_{\psi}(s)] + \mathbb{E}_{s \sim \pi_A}[\log(1 - D_{\psi}(s))]
    \label{eq:gail}
\end{align}
where $\pi_A$ is the agent and $\pi_E$ the expert policy.
The discriminator can be used to obtain a reward function, which in our case is simply $R(s) = -\log(1 - D_{\psi}(s))$.
The policy $\pi_A$ is trained to maximize this reward function.
The original GAIL~\cite{ho2016generative} was trained on-policy with an additional entropy regularizer, but has since been adapted to the off-policy actor-critic setting~\cite{kostrikov2018discriminator,reed2018visual}.

\section{Task-Relevant Adversarial Imitation Learning (TRAIL)}
%
\iffalse
\Huge
\definecolor{Ie}{rgb}{0.21960784313, 0.46274509803, 0.11372549019}
\textcolor{Ie}{$\bm{\mathcal{I}_E}$}
%
\definecolor{Ia}{rgb}{0.6, 0.0, 0.0}
\textcolor{Ia}{$\bm{\mathcal{I}_A}$}
\fi
%
\begin{figure}[t]
  \vspace{-0.1in}
  \centering
  \includegraphics[width=0.45\linewidth]{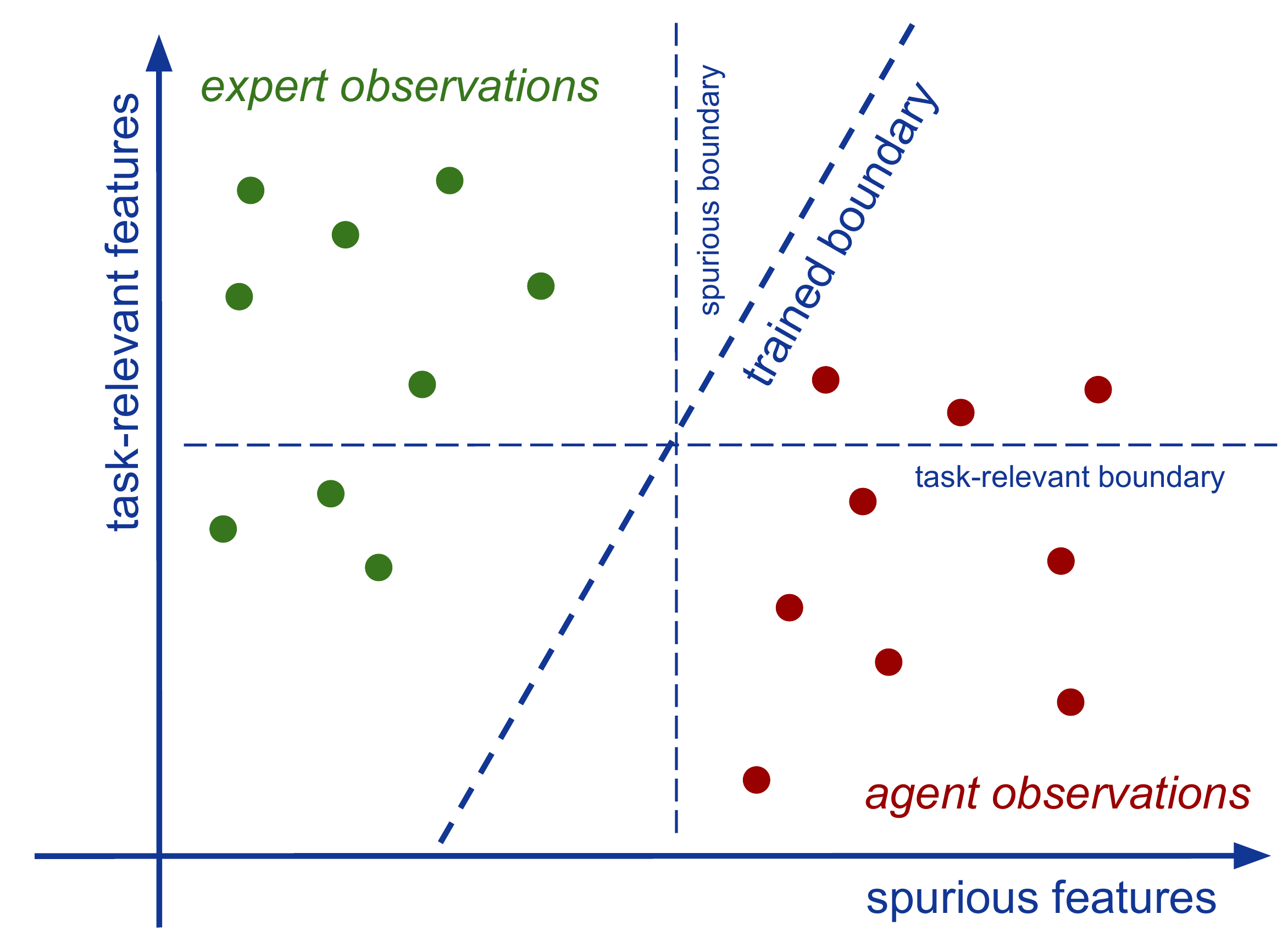}
  \hspace{0.2cm}
  \includegraphics[width=0.45\linewidth]{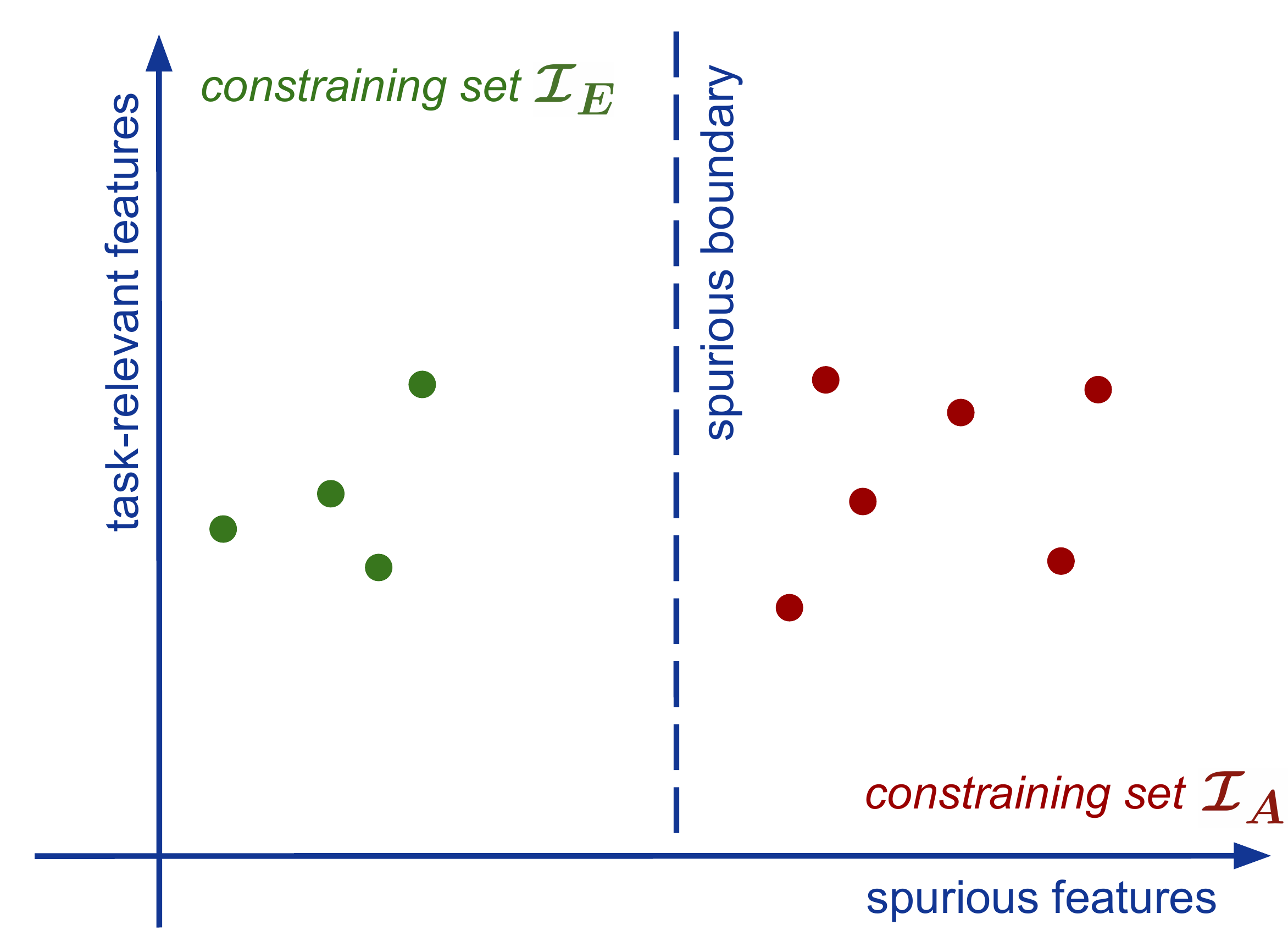}
  \caption{On the left, we illustrate a trained boundary separated into spurious and task-relevant components.
  Ideally, to provide informative rewards, our discriminator should only consist of the task-relevant component.
  On the right, constraining sets $\mathcal{I}_E$ and $\mathcal{I}_A$ are constructed such that only spurious features can be used to discriminate between them, which isolates the spurious decision boundary. Intuitively, our method works by \emph{unlearning} this spurious boundary, so that the discriminator better captures the task-relevant boundary.}
  \label{fig:two_boundaries}
  \vspace{-0.05in}
\end{figure}
TRAIL is designed to prevent the discriminator from forming spurious associations.
To this end, we regularize the discriminator to be unable to distinguish between \emph{constraining sets} $\mathcal{I}_E$ and $\mathcal{I}_A$, which consist of expert and agent observations, respectively, such that their elements can \emph{only} be identified as belonging to one or the other set using spurious features.
See Figure~\ref{fig:two_boundaries} for an illustration.

The TRAIL discriminator uses a cross-entropy objective as in GAIL (Eq.~\ref{eq:gail}), but with an accuracy constraint applied to observations from $\mathcal{I}_E \cup \mathcal{I}_A$.
For these observations, the cross-entropy objective is reversed if
\mbox{$accuracy(\mathcal{I}_E, \mathcal{I}_A) \geq \frac{1}{2}$}
and zero otherwise, where 
$accuracy(\mathcal{I}_E, \mathcal{I}_A)$ is defined as the average discriminator accuracy on a balanced set of observations from the constraining sets, i.e.
\begin{equation}
\frac{1}{2} \mathbb{E}_{s \in \mathcal{I}_E}\left[ \1_\mathrm{D_\psi(s) \geq \frac{1}{2}}\right] + \frac{1}{2} \mathbb{E}_{s \in \mathcal{I}_A}\left[ \1_\mathrm{D_\psi(s)< \frac{1}{2}}\right].
\end{equation}
Intuitively, we discourage the discriminator from identifying and exploiting spurious patterns, and force the discriminator to forget them if they are in use.

Concretely given a batch of $N$ examples $s_E \sim \pi_E$, $s_A \sim \pi_A$ from the expert and agent, and constraining observations $\hat{s}_E \subset \mathcal{I}_E$ and $\hat{s}_A \subset \mathcal{I}_A$, the TRAIL discriminator maximizes
\begin{align}
\mathcal{L}_{\psi}(s_E, s_A, \hat{s}_E, \hat{s}_A) = & \quad G_\psi(s_E, s_A) - \1_{accuracy(\hat{s}_E, \hat{s}_A) \geq \frac{1}{2}} G_\psi(\hat{s}_E, \hat{s}_A), \nonumber
\end{align}
where $G_\psi(s_E, s_A)$ is finite sample estimate of the GAIL discriminator loss for $D_\psi$, which equals
\begin{equation}
G_\psi(s_E, s_A) = \sum_{i=1}^{N}\log D_\psi(s_E^{(i)}) + \log[1 - D_\psi(s_A^{(i)}) ],
\end{equation}
and $\1_{accuracy(\hat{s}_E, \hat{s}_A)\geq \frac{1}{2}}$ indicates whether the constraint is violated.

\textbf{Constraining sets}\hspace{0.2cm}
$\mathcal{I}_E$ and $\mathcal{I}_A$ are drawn from demonstrations and agent episodes, respectively, such that an observation can \emph{only} be identified as belonging to one or the other using spurious features.

In general, tasks with non-stationary distributional shifts could be devised, which would make it difficult or impossible to construct such sets.
However, in many situations of interest, including our robotic manipulation setup, it is easy to propose effective and general constraining sets.

For example, a straightforward way to collect observations suitable for $\mathcal{I}_E$ and $\mathcal{I}_A$ is to execute a random policy in both expert and agent settings.
This yields observations of purposeless behaviour, with no task information, varying only in task-irrelevant features.
Another way to construct $\mathcal{I}_E$ and $\mathcal{I}_A$ is to use \emph{early frames} from expert and agent episodes.
Since in early frames little or no task behavior is present, this strategy turns out to be effective and no extra data has to be collected.
This strategy also improves robustness with respect to variation in the initial conditions of the task; see for example block insertion in Figure~\ref{fig:trail_tasks}(d).

Importantly, if the sets $\mathcal{I}_E$ and $\mathcal{I}_A$ capture some type of irrelevance but not all types, their inclusion will still result in improvements in performance.
In this regard, TRAIL improves over GAIL whenever the designer has any prior knowledge of which aspects of the data are task-irrelevant.

\section{Experimental setup}\label{sec:exp_setup}

\begin{wrapfigure}{r}{0.5\textwidth}
  \centering
  \vspace{-0.5in}
  \includegraphics[width=0.48\linewidth,height=0.45\linewidth]{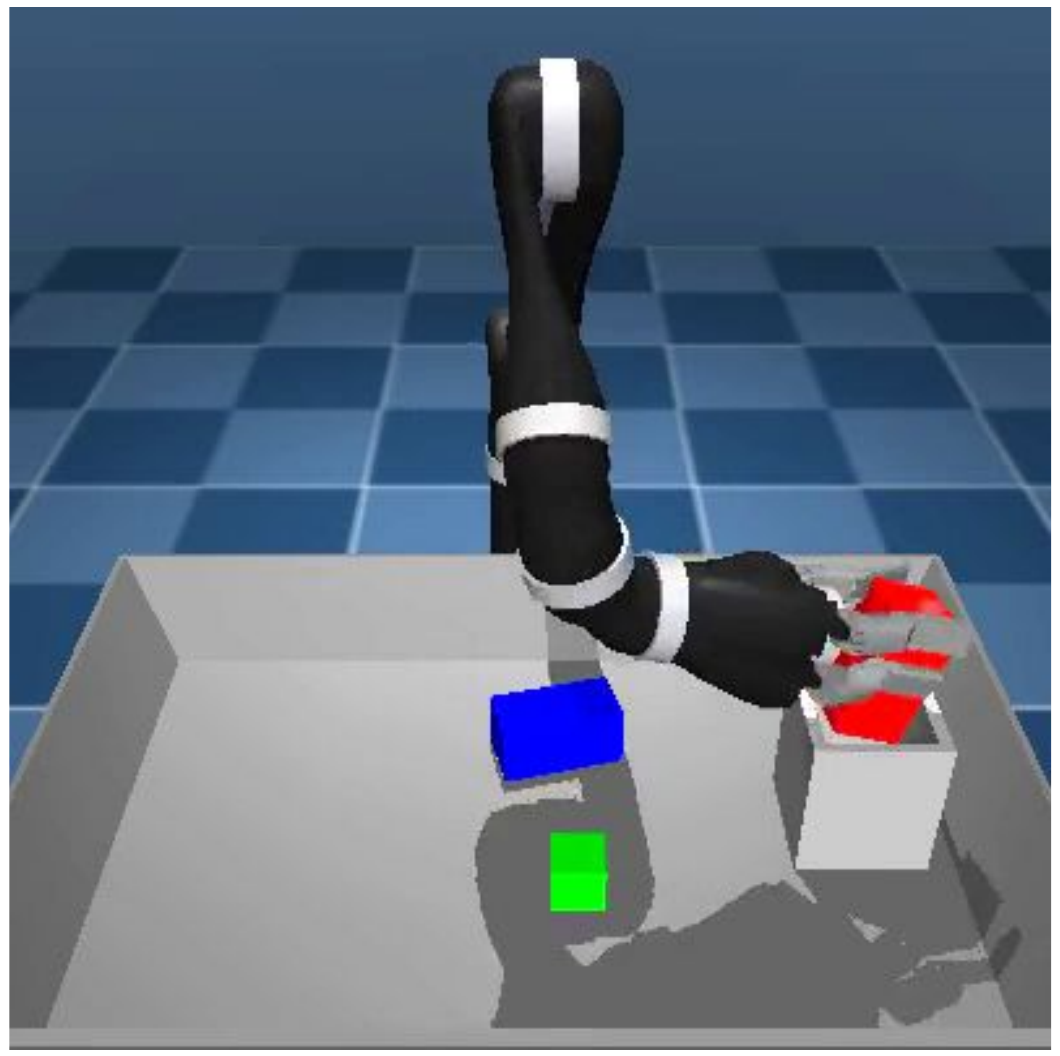}\vspace{0.05cm}\hspace{0.01cm}
  \includegraphics[width=0.48\linewidth,height=0.45\linewidth]{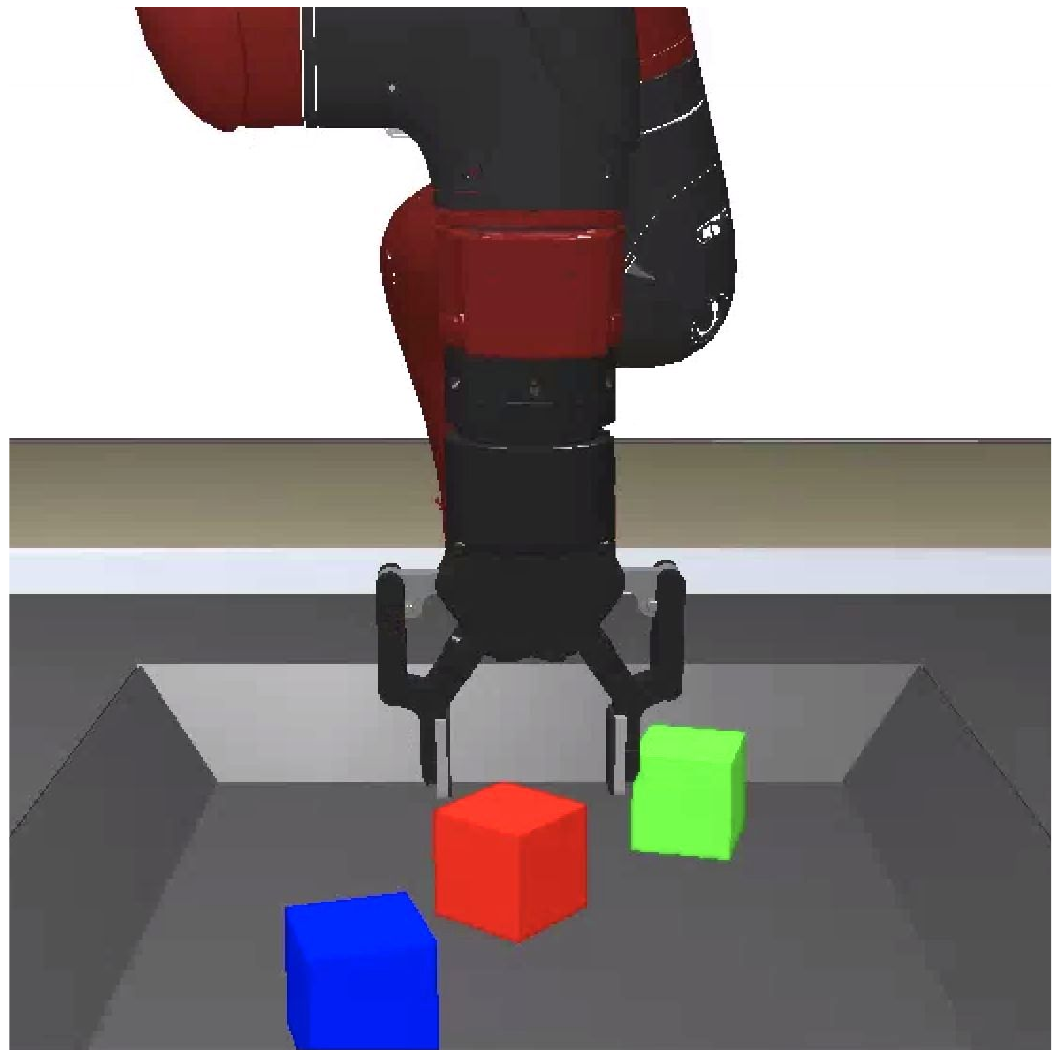}
  \caption{Two work spaces, \textit{Jaco} (left) which uses the Jaco arm and is 20 $\times$ 20 cm, and \textit{Sawyer} (right) which uses the Sawyer arm and more closely resembles a real robot cage and is 35 $\times$ 35 cm.}
  \label{fig:cage_env_main}
  \vspace{-0.2in}
\end{wrapfigure}
\textbf{Environments}\hspace{0.2cm}
We focus on solving simulated robot manipulation tasks.
The environment implements two work-spaces: one with a Kinova Jaco arm (\textit{Jaco}), and the other with a Sawyer arm (\textit{Sawyer}).
See Figure~\ref{fig:cage_env_main} for visualization and supplementary material~\ref{sec:env_details} for a detailed description.
Environment task rewards (which are not required by GAIL or by TRAIL) are sparse and equal to +1 for each step when a given task is solved and 0 otherwise.
The maximum reward for the episode is 200, because this is the length of a single evaluation episode.
For each task we collect 100 human demonstrations.

\textbf{Data augmentation}\hspace{0.2cm}
We find that data augmentation is necessary to prevent discriminator overfitting and therefore, it is used in all discriminator-based methods in this work (including all baselines).
See supplementary material~\ref{sec:data_augmentation} for details and relevant ablations.

\textbf{Actor early stopping}\hspace{0.2cm}
When the agent starts to act as desired, it generates data that is indistinguishable from the expert in its behaviour.
This forces the discriminator to rely on spurious features.
To avoid this, we propose to stop and restart each actor episode after a certain number of steps such that successful behavior is rarely represented in agent data.
This enables the discriminator to recognize the goal condition as representative of expert behavior, because it now appears much more frequently in expert observations than in agent ones.
The optimal stopping step number can be found using grid search but to avoid hand-tuning, we found that the discriminator score can be used to derive an adaptive stopping criterion.
Concretely, we restart an episode if the discriminator score at the current step exceeds the median score of the episode so far for $T_{patience}$ consecutive steps (in practice we set $T_{patience}=10$).
In all experiment we use the adaptive method unless clearly stated in the text. See ablations in Section~\ref{sec:ablations} for comparison.

\textbf{TRAIL}\hspace{0.2cm}
Our method proposes to modify the discriminator objective.
This is orthogonal to the rest of the choices that constitute GAIL algorithms (for example on-policy or off-policy).
We keep these choices the same to guarantee fair comparison across methods.
We used early frames to construct the constraining sets except as indicated in the relevant ablations (see Section~\ref{sec:constructing}).

\textbf{Baselines}\hspace{0.2cm}
The most fundamental baselines are behavior cloning, which can be trained from demonstrations only (without any interactions with the environment), and conventional GAIL.
The baseline GAIL in our experiments uses Eq.~\ref{eq:gail} to train its discriminator.
It is comparable to our proposed TRAIL, only differing in the discriminator objective function (due to the use of constraining sets) and actor early stopping.
We also implement \aes{} which is the baseline GAIL enhanced with actor early stopping.
Therefore, the only difference between \aes{} and TRAIL, is that the latter uses constraining sets. This baseline helps us to evaluate the significance of all components.
Our RL agents are based on the off-policy actor-critic D4PG algorithm \citep{barth2018distributed} because of its stability and data-efficiency (see supplementary material~\ref{appendix:d4pg}).
Following \citet{vecerik2017learning}, we add expert demonstrations into the agents' experience replay, and refer to the resulting RL algorithm as D4PG from Demonstrations (D4PGfD).
The GAIL-based models in this paper differ from D4PGfD only by the use of the discriminator-based rewards instead of ground truth environment rewards.

We used single frames as observations but sequences could also be used, and we do not require expert actions.
Not requiring actions enables learning in very off-policy settings, where the action dimensions and distributions of the demonstrator (another robot or human) could be different from those of the agent \citep{zolna2019reinforced}.
Finally, to make our comparisons as fair as possible we used a shared implementation, and the same agent configurations (e.g. number of actors) and network architecture.

\section{Results}

\subsection{Evaluation on diverse manipulation tasks}
We first focus on four tasks which do not have any purposely introduced spurious features; \emph{lift} and \emph{stack} with the \emph{Sawyer} robot, and \emph{stack banana} and \emph{insertion} with the \emph{Jaco} robot.

\begin{figure}[ht]
  \centering
  \includegraphics[height=0.19\linewidth]{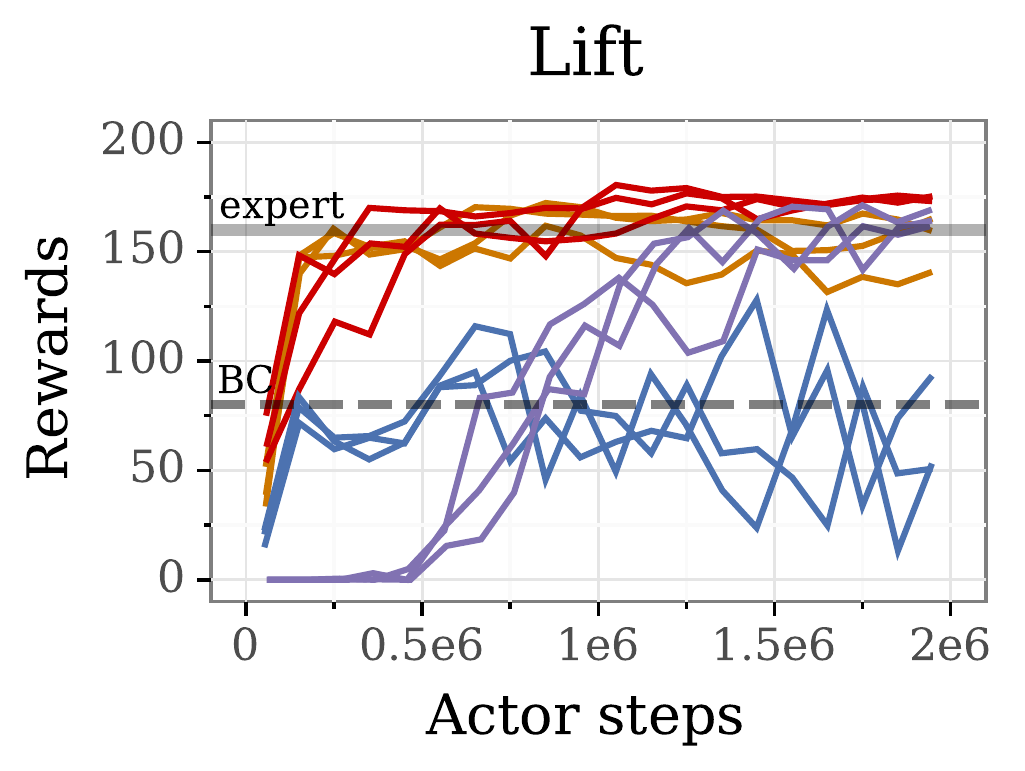}
  \hspace{-0.15cm}
  \includegraphics[height=0.19\linewidth]{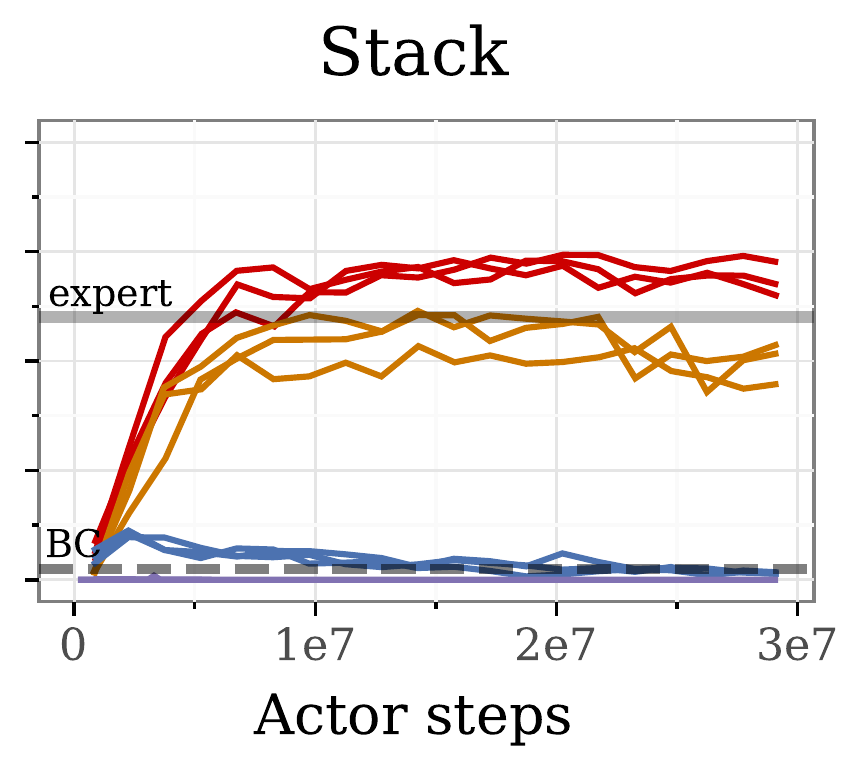}
  \hspace{-0.15cm}
  \includegraphics[height=0.19\linewidth]{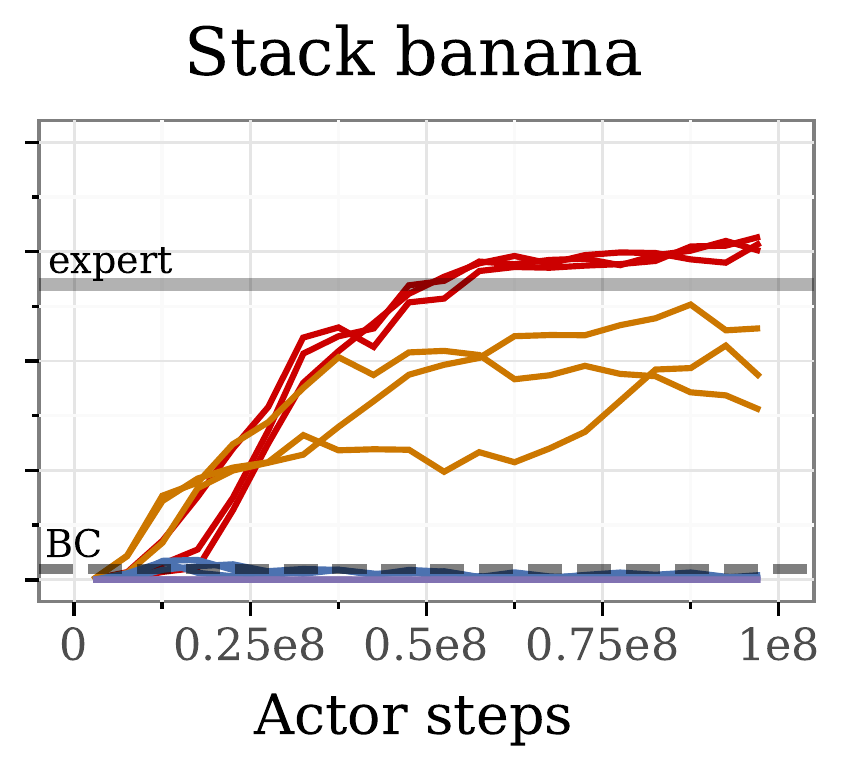} \hspace{-0.15cm}
  \includegraphics[height=0.19\linewidth]{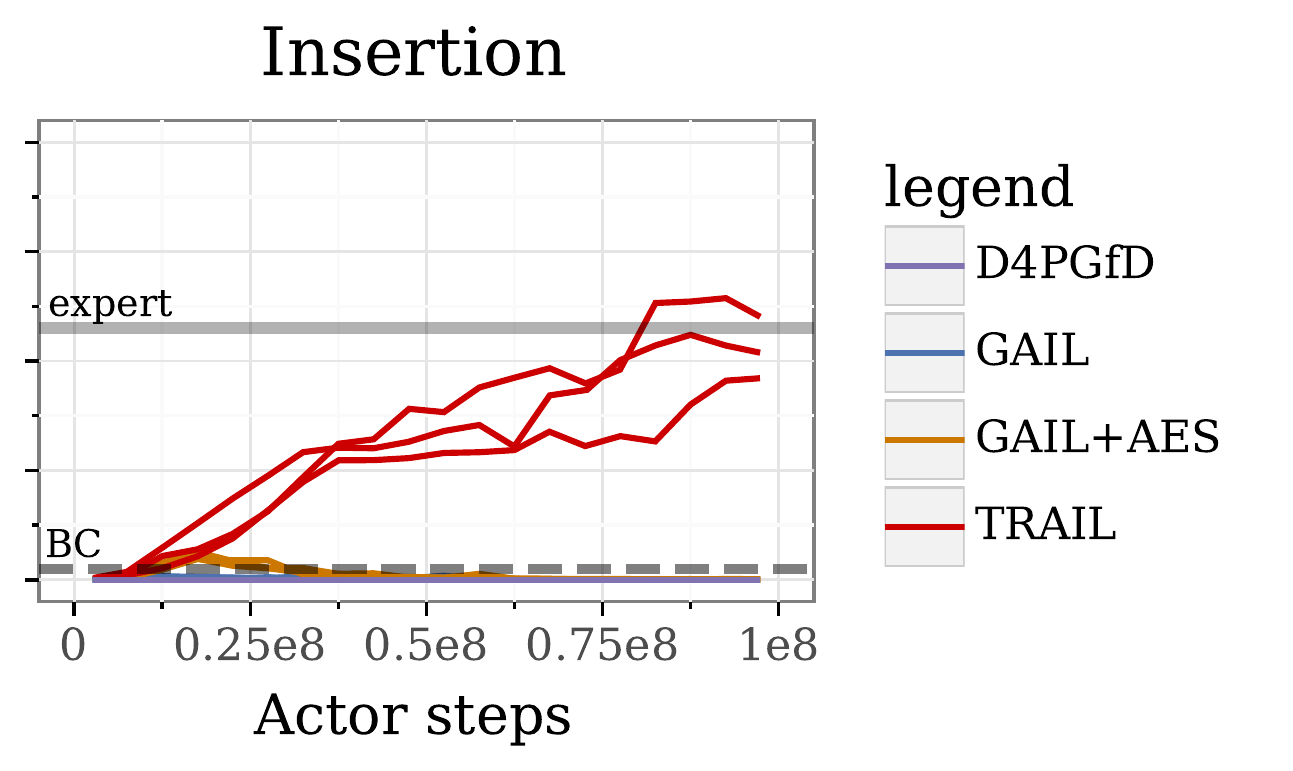}
  \caption{Results comparing TRAIL, \aes{} and baselines for diverse manipulation tasks.}
  \label{fig:extra_tasks}
\end{figure}

The results are shown in Figure~\ref{fig:extra_tasks}.
The tasks are very challenging as evidenced by the performance of BC agents which only partially solve one task -- \emph{lift}.
The baseline GAIL fails, even though the experiments are conducted in fully controlled environments (seemingly free of spurious features), indicating that GAIL discriminators identify less evident spurious features like positions of distracting objects in expert demonstrations.
TRAIL reaches expert-level performance on all four tasks.
\aes{} always performs worse than TRAIL and completely fails for the \emph{insertion} task, stressing the importance of the constraining sets which directly address the problem of spurious associations.

D4PGfD, which uses sparse ground truth rewards, is able to solve only \emph{lift}, which suggests that dense discriminator-based rewards aid exploration and are critical to solve more challenging tasks.

\subsection{Expert with different appearance}

\begin{figure}[h]
  \centering
  \includegraphics[height=0.225\linewidth]{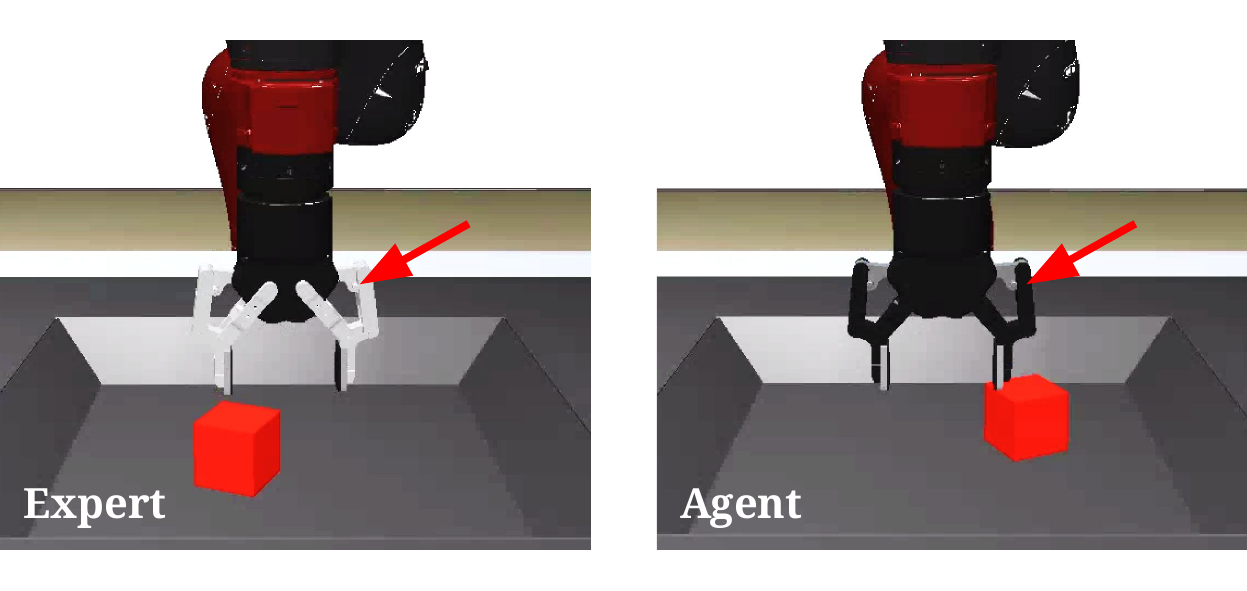}
  \hspace{0.2cm}
  \includegraphics[height=0.225\linewidth]{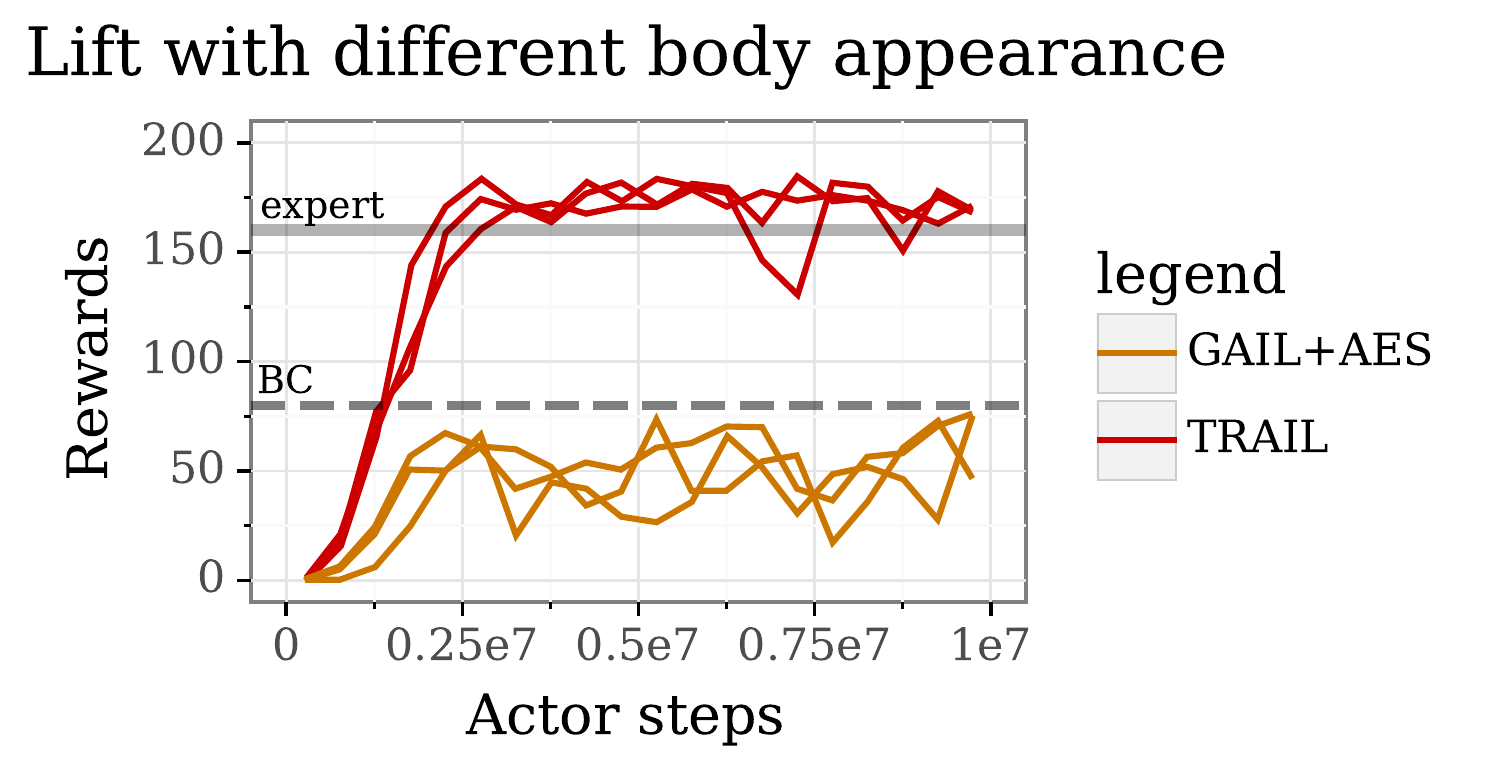}
  \caption{
  When the appearance differs (as highlighted by the arrow), TRAIL clearly outperforms baselines.\label{fig:different_body}}
\end{figure}

In this section, we intentionally introduce a visual spurious feature by changing the expert appearance, increasing the difficulty of the \emph{lift} task.
Figure~\ref{fig:different_body} shows the difference in appearance, which is to vary the gripper color from light to dark.
This is meant to simulate the type of distributional shifts that commonly occur when dealing with real robots, where the appearance can change after initial data collection due to scratches and wear, for example.
The results are presented in Figure~\ref{fig:different_body}.

With the modified expert appearance, \aes{} performance degrades in block lifting, despite the use of data augmentation and the addition of 100-step actor early stopping (AES).
TRAIL solves the task and achieves performance better than the expert thus being robust to spurious features.

\subsection{Block lifting with distractors}\label{sec:baselines}
The previous section showed that a consistent difference in expert appearance degrades the performance of GAIL, while TRAIL remains unaffected. 
In this section, \emph{instead} of changing the expert appearance, we alter the positions of props in the environment.
We consider another variant of the lift task, \textit{lift distracted}, when two extra blocks are added (blue and green, see Figure~\ref{fig:trail_tasks}(a,b)).
Here, there are \emph{no} differences in expert appearance.

In this section, on top of the previously defined baselines, we consider additional GAIL-based approaches proposed by \citet{reed2018visual}; using either a randomly initialized convolutional network, or a convolutional critic network, to provide fixed vision features on top of which a tiny discriminator network is trained. We call these two baselines GAIL+random and GAIL+critic respectively.

\begin{figure}[ht]
  \centering
  \includegraphics[height=0.225\linewidth]{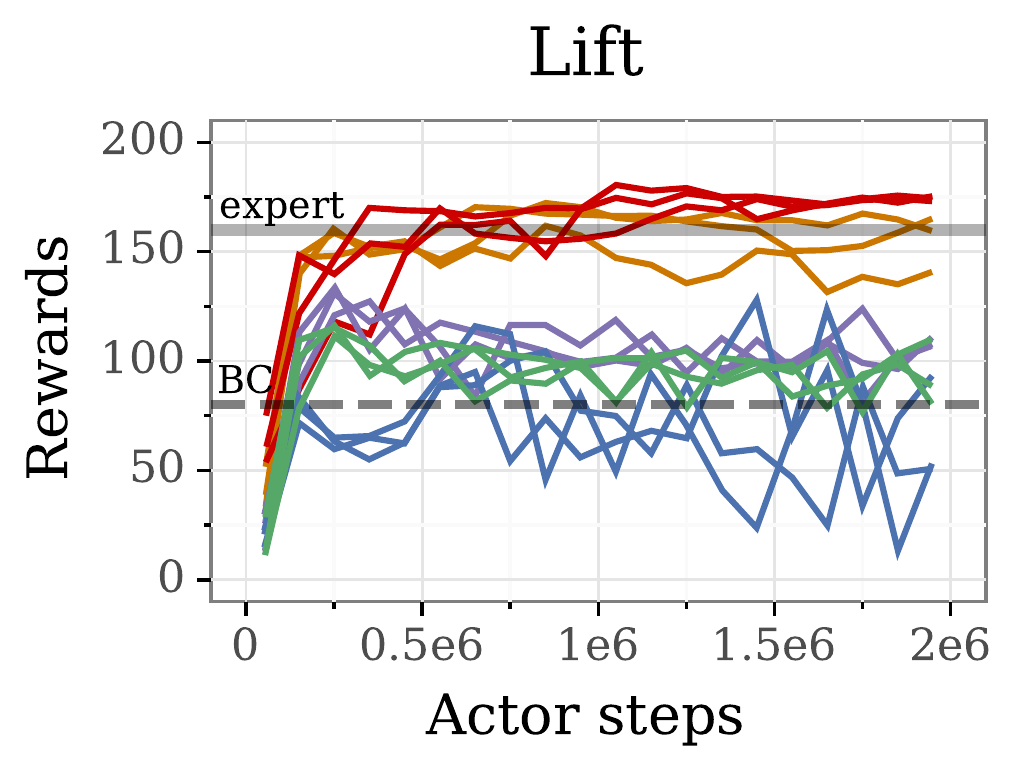} \hspace{-0.15cm}
  \includegraphics[height=0.225\linewidth]{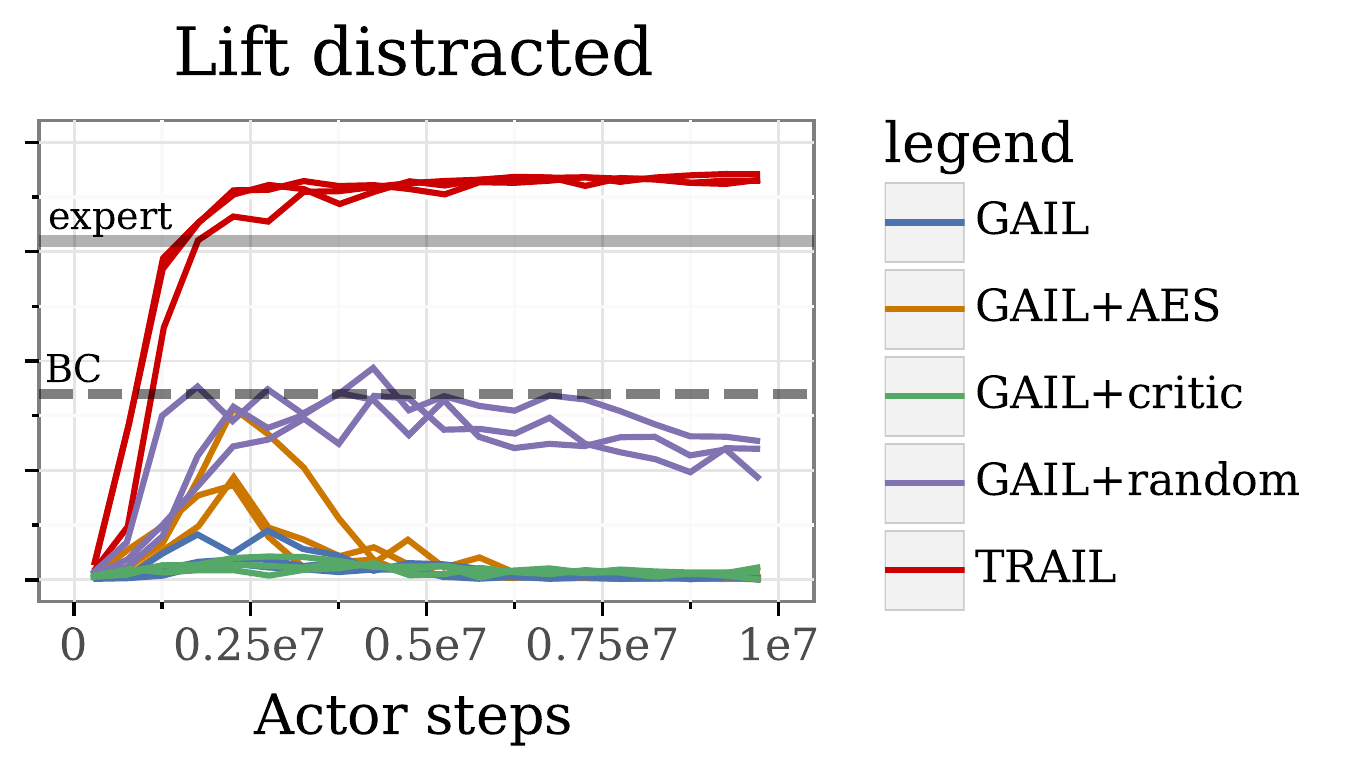}
  \caption{Results for \textit{lift} and \textit{lift distracted}. When distractors are introduced only TRAIL succeeds.
  \label{fig:rgb_lift}}
  \vspace{-0.05in}
\end{figure}

As shown in Figure~\ref{fig:rgb_lift}, all methods, including new baselines, perform satisfactorily on \textit{lift}, but \aes{} and TRAIL clearly do best.
As expected, the performance of BC on \textit{lift distracted} is similar to its performance on \textit{lift}, despite the two additional blocks.
The two additional blocks, however, affect all GAIL-based baselines, despite being irrelevant to the task, while TRAIL succeeds.

Even for this relatively easy task in the simulated environment without evident task-irrelevant features, the baseline discriminators find spurious associations to obtain perfect accuracy.
It is achieved by simply memorizing all 100 initial block positions from the demonstration set.
Note that this can not be prevented using unstructured regularization, as a discriminator has to be able to infer blocks' positions from the image to provide informative reward.
However, the positions of task-irrelevant objects should be ignored.
TRAIL is able to handle the variety of initial cube positions, achieving better than expert performance on \textit{lift distracted}.

We conducted an additional experiment termed \textit{lift distracted seeded}, in which the initial block positions are randomly drawn from the expert demonstrations.
Therefore, it is impossible to reliably discriminate between expert and agent episodes using distractor positions.
\aes{} performed much better on \textit{lift distracted seeded} than on \textit{lift distracted}, indicating that the discriminator does form a spurious association between distractor positions and expert labeling.
We provide more detailed analysis of \textit{lift distracted} (and the \textit{lift distracted seeded} variant) in supplementary material~\ref{sec:lift_distracted_abl}.

\subsection{Constructing the constraining sets $\mathcal{I}_E$ and $\mathcal{I}_A$}\label{sec:constructing}

\begin{figure}[h]
  \vspace{0.05in}
  \centering
  \includegraphics[height=0.225\linewidth]{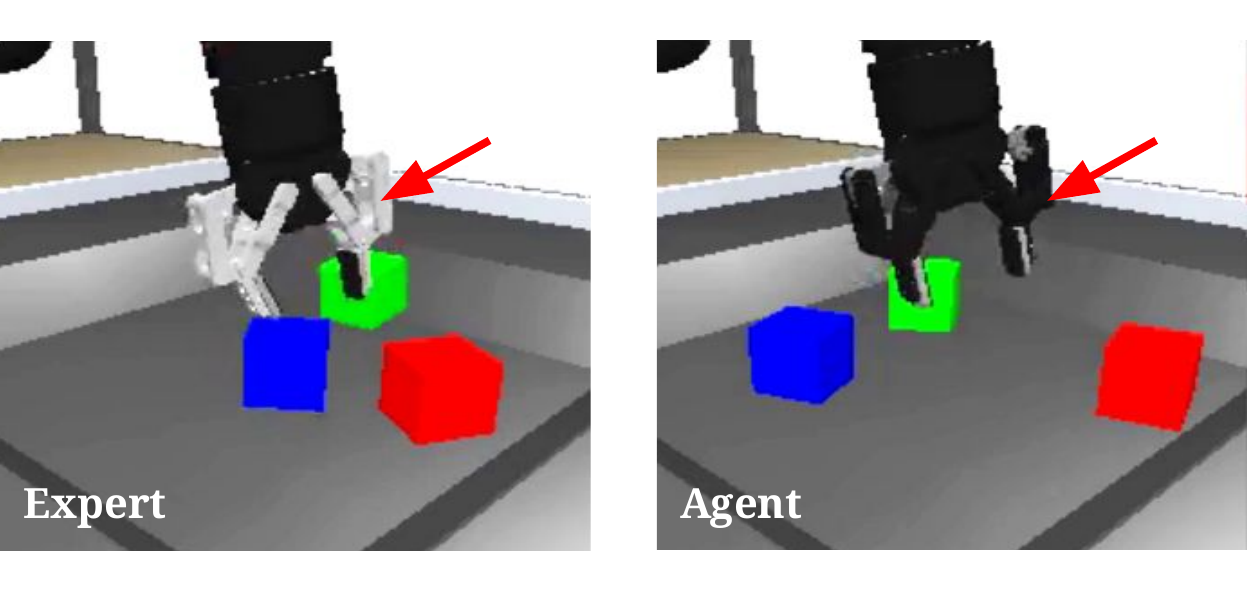}
  \hspace{-0.2cm}
  \includegraphics[height=0.225\linewidth]{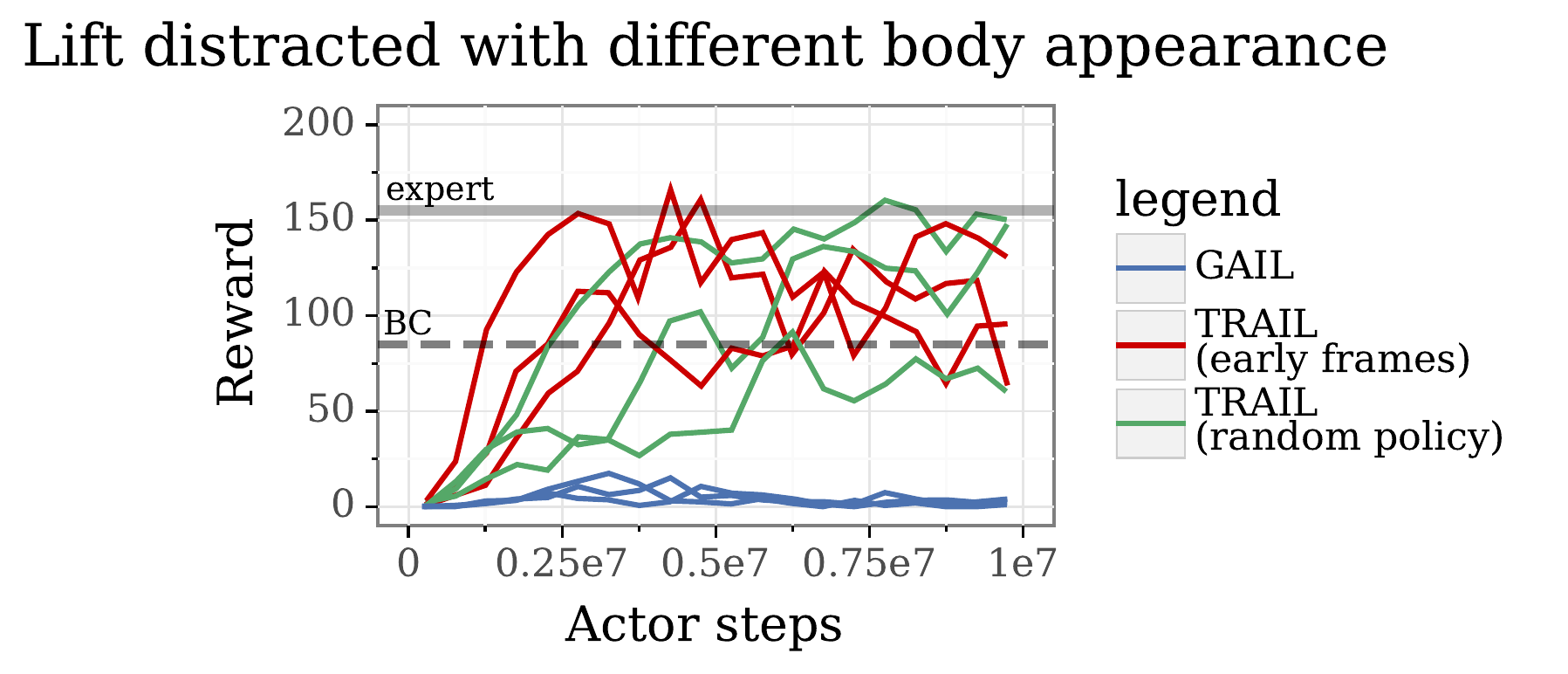}
  \caption{
  The hardest variant of \emph{lift} (red cube) task; not only robot appearance changes but also distractors are added. There is no clear difference between TRAIL variants but both of them decisively outperform GAIL. \label{fig:different_body_distractors}}
\end{figure}

In the previous sections, early frames were used to construct the constraining sets.
Here, we compare the performance of this approach with another previously mentioned strategy: the use of a random policy to construct $\mathcal{I}_E$ and $\mathcal{I}_A$.
To tease out the differences between these two variants, we consider a harder task by combining two previous tasks: \emph{lift distracted} and \emph{different body appearance}.
The results are presented in Figure~\ref{fig:different_body_distractors} which also shows the difference in robot appearance and distractors.

The new task is indeed harder and it takes longer for TRAIL methods to achieve performance better than the BC baseline, which is not affected by the different body appearance. GAIL is clearly outperformed and does not take off.

The performance of TRAIL-early and TRAIL-random is similar.
Hence, by default we simply use early frames. This choice is pragmatic as it does not require that we collect any additional data, and hence the comparison with GAIL and other baselines is fair. 
It is also simple to apply in practice, even if one no longer has access to the expert setup.
Finally, it is general enough to be successfully used across all robotic manipulation tasks considered in this work.

To decide how many initial frames should be used to construct $\mathcal{I}_E$ and $\mathcal{I}_A$, we conducted an ablation study and found that the method is not very sensitive to this choice (see supplementary material~\ref{sec:early_frames}).
Hence, we chose 10 initial frames, and intentionally used the same number for all tasks to further emphasize generality of this choice.

\subsection{Ablation studies}\label{sec:ablations}
\textbf{Actor early stopping}\hspace{0.2cm}
\label{sec:experiments_early_stopping}
We analyze the importance of actor early stopping on 3 tasks in the \textit{Jaco} work-space: lift red cube (\textit{lift}), put red cube in box (\textit{box}), and stack red cube on blue cube (\textit{stack}).
We consider three variants of \aes{} with the following termination policies: a) fixed step~(50), b) adaptive (as described in Section~\ref{sec:exp_setup}), and c) based on ground truth task rewards (as adaptive but uses ground truth rewards instead of discriminator estimates; we also use $T_{patience}=10$).

\begin{figure}[h]
  \centering
  \includegraphics[height=0.225\linewidth]{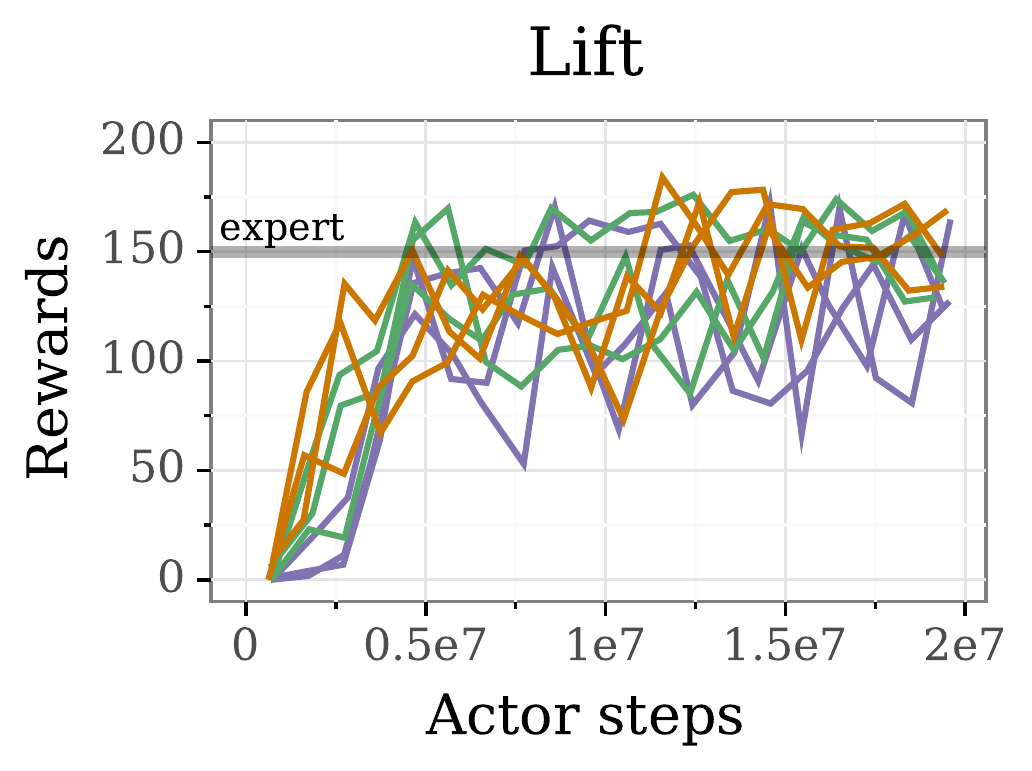} \hspace{-0.2cm}
  \includegraphics[height=0.225\linewidth]{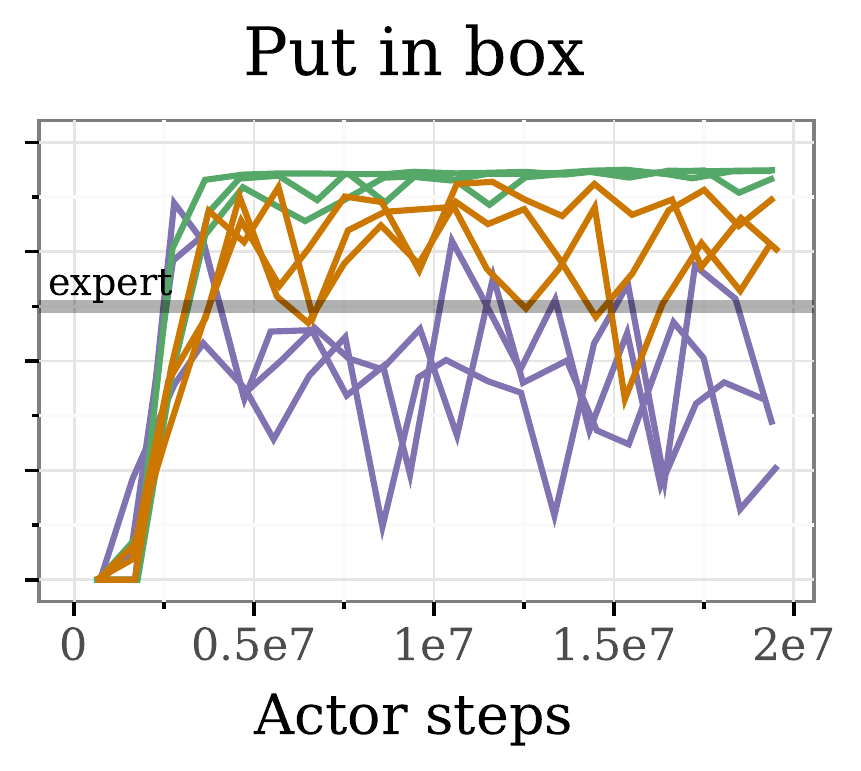} \hspace{-0.2cm}
  \includegraphics[height=0.225\linewidth]{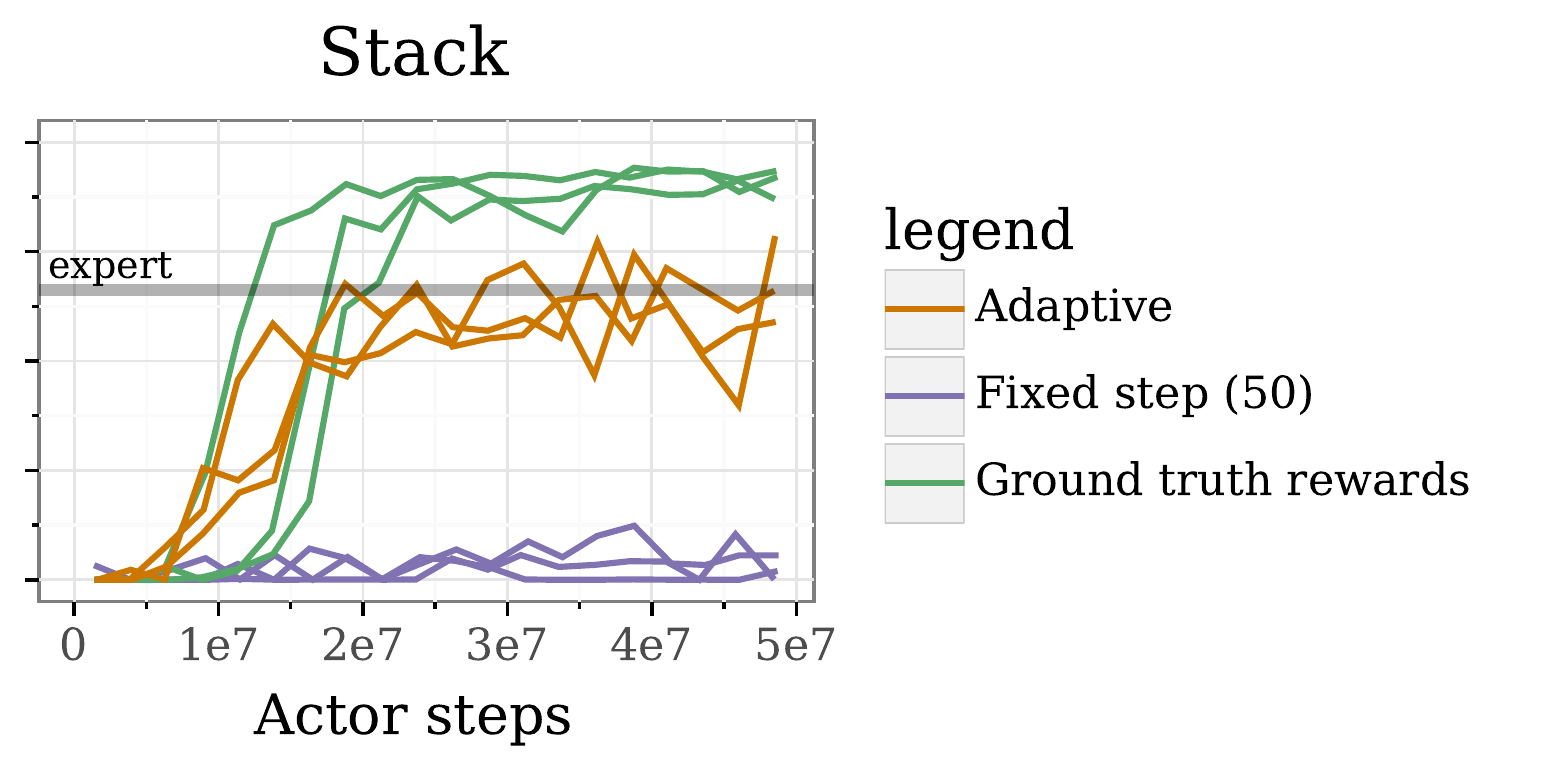}
  \caption{Results for \textit{lift}, \textit{box}, and \textit{stack} on \textit{Jaco} environment for \aes{} with different variants of AES.}
  \label{fig:easy-tasks}
\end{figure}

Results are presented in Figure~\ref{fig:easy-tasks}.
Termination based on task reward is clearly superior; although unrealistic in practice, it defines the performance upper-bound and clearly shows that actor early stopping is beneficial. 
The adaptive approach is robust and reaches human performance on all tasks. A fixed termination policy, when tuned, can be very effective. The same fixed termination step, however, does not work for all tasks (see supplementary material~\ref{sec:fixed_termination} for further details).

The experiments show that successful agent episodes can trigger formation of spurious associations and this can be prevented directly with actor early stopping.
However, the method is not sufficient in more difficult cases, as showed in the previous experiments on more challenging tasks.

\begin{wrapfigure}{r}{0.5\textwidth}
  \vspace{-0.2in}
  \centering
  \includegraphics[height=0.45\linewidth]{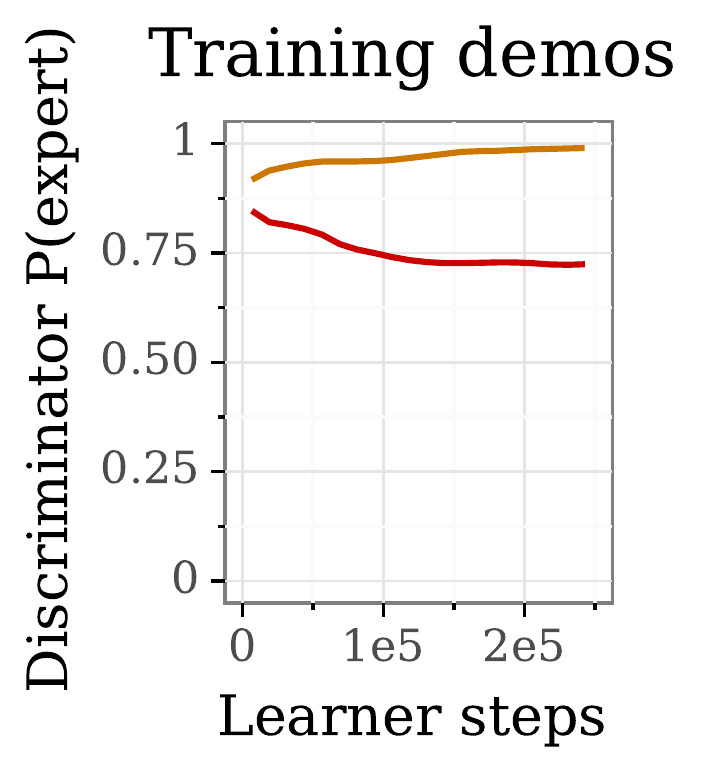}
  \hspace{-0.2cm}
  \includegraphics[height=0.45\linewidth]{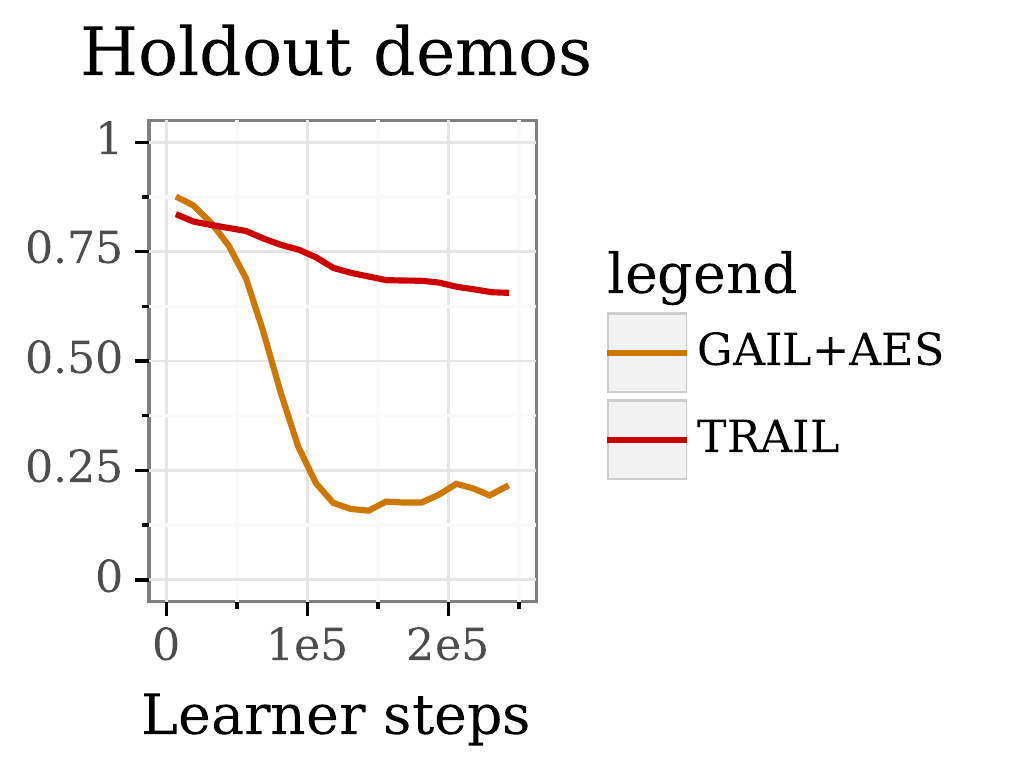}
  \caption{TRAIL generalizes better than \aes{}.
  \label{fig:generalization}}
  \vspace{-0.15in}
\end{wrapfigure}
\textbf{TRAIL generalization abilities}\hspace{0.2cm}
We analyze how the use of constraining sets affects discriminator generalization.
We compare TRAIL with \aes{}, as the only difference between them is the use of constraining sets.
We use the \textit{lift distracted} task.
To measure the discriminator's generalization capacity, we collected 25 extra holdout demonstrations.
We visualize and compare discriminator predictions on training and hold-out demonstrations in Figure~\ref{fig:generalization}.

The TRAIL discriminator generalizes well, as it assigns similar values to training and hold-out demos throughout the training (red curves are aligned to each other).
The \aes{} discriminator quickly forms spurious associations which do not generalize to holdout demos.
Its predictions for training demos approach $1$ while holdout demos are recognized as agent trajectories (predictions below $0.5$). 

\textbf{Learning with an oracle discriminator}\hspace{0.2cm}
To further understand the contribution of learned reward functions, we train agents with rewards from an oracle discriminator, which always assigns a reward of $1$ for expert frames and $0$ for agent frames.
This simulates a discriminator that perfectly exploits a spurious association between task-irrelevant visual features and expert labels.
On the \textit{lift} task, agents using this fixed oracle reward achieve roughly half the reward of TRAIL asymptotically, and on \textit{lift distracted} they do not take off.
See supplementary material~\ref{sec:oracle_discriminator} for learning curves.

\section{Conclusions}

We have shown that a critical vulnerability in GAIL is the tendency of discriminators to form spurious associations between visual features and expert labels, which results in uninformative rewards.
The problem can be triggered by many factors which can be addressed directly to improve agent performance, as done by \aes{}, but it does not scale to harder problems.
TRAIL, however, does not require any knowledge about potential spurious associations as they are neutralized automatically.
Without access to expert actions nor environment rewards, TRAIL achieved expert level on a diverse set of manipulation tasks, outperforming comparable GAIL, behavior cloning, and D4PGfD agents.

% The maximum paper length is 8 pages excluding references and acknowledgements, and 10 pages including references and acknowledgements

\clearpage
% The acknowledgments are automatically included only in the final and preprint versions of the paper.

\acknowledgments{
We would like to thank reviewers for their useful comments. Konrad \.Zo\l{}na is supported by the National Science Center, Poland (\mbox{2017/27/N/ST6/00828}, \mbox{2018/28/T/ST6/00211}).
}

\end{bibunit}

\begin{bibunit}

\clearpage
\appendix
{\Large \bf Supplementary material}
\normalsize
\label{sec:sup_mat}

\section{Detailed description of environment}
\label{sec:env_details}

\begin{figure}[ht]
  \centering
  \vspace{0.1cm}
  \includegraphics[width=0.225\linewidth,height=0.225\linewidth]{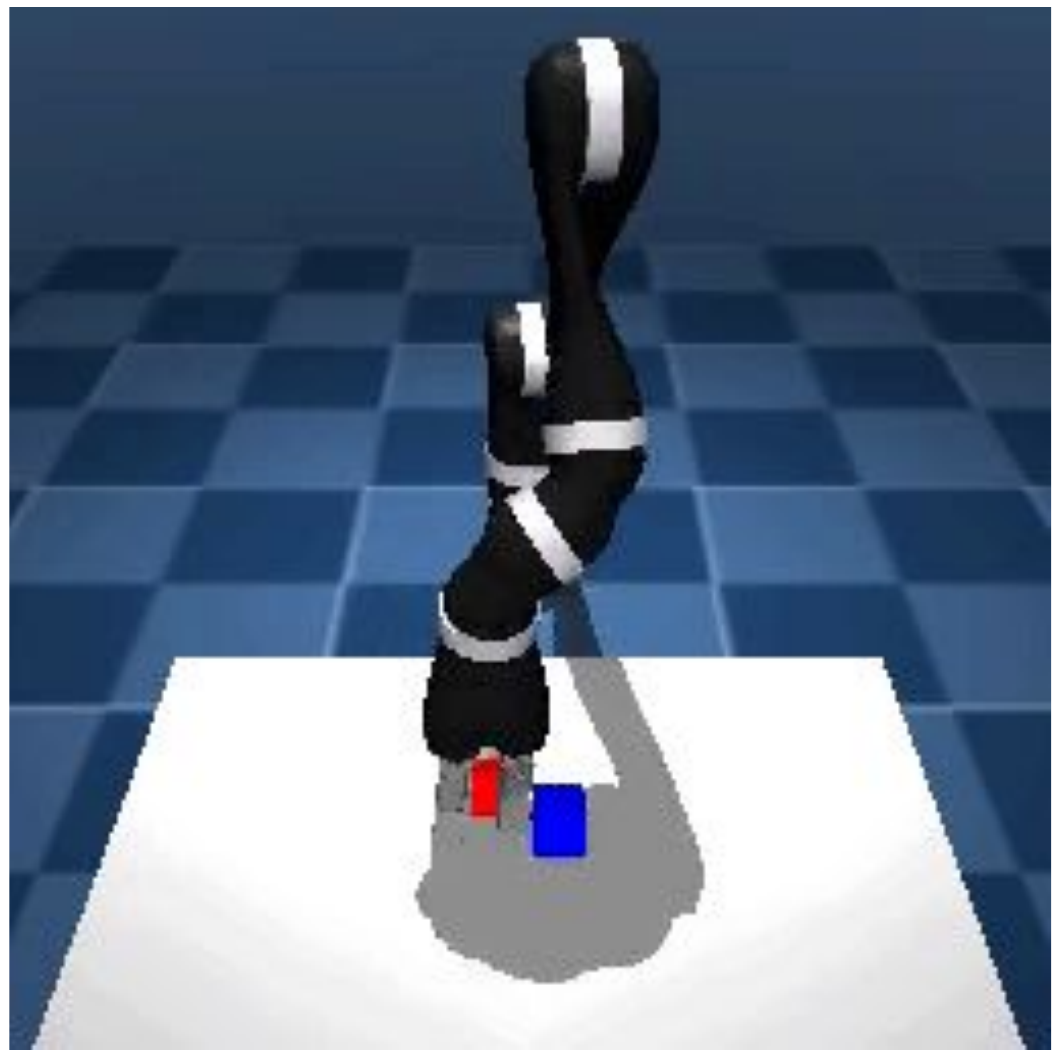}
  \includegraphics[width=0.225\linewidth,height=0.225\linewidth]{insertion.pdf}
  \vspace{0.05cm}\hspace{0.01cm}
  \includegraphics[width=0.225\linewidth,height=0.225\linewidth]{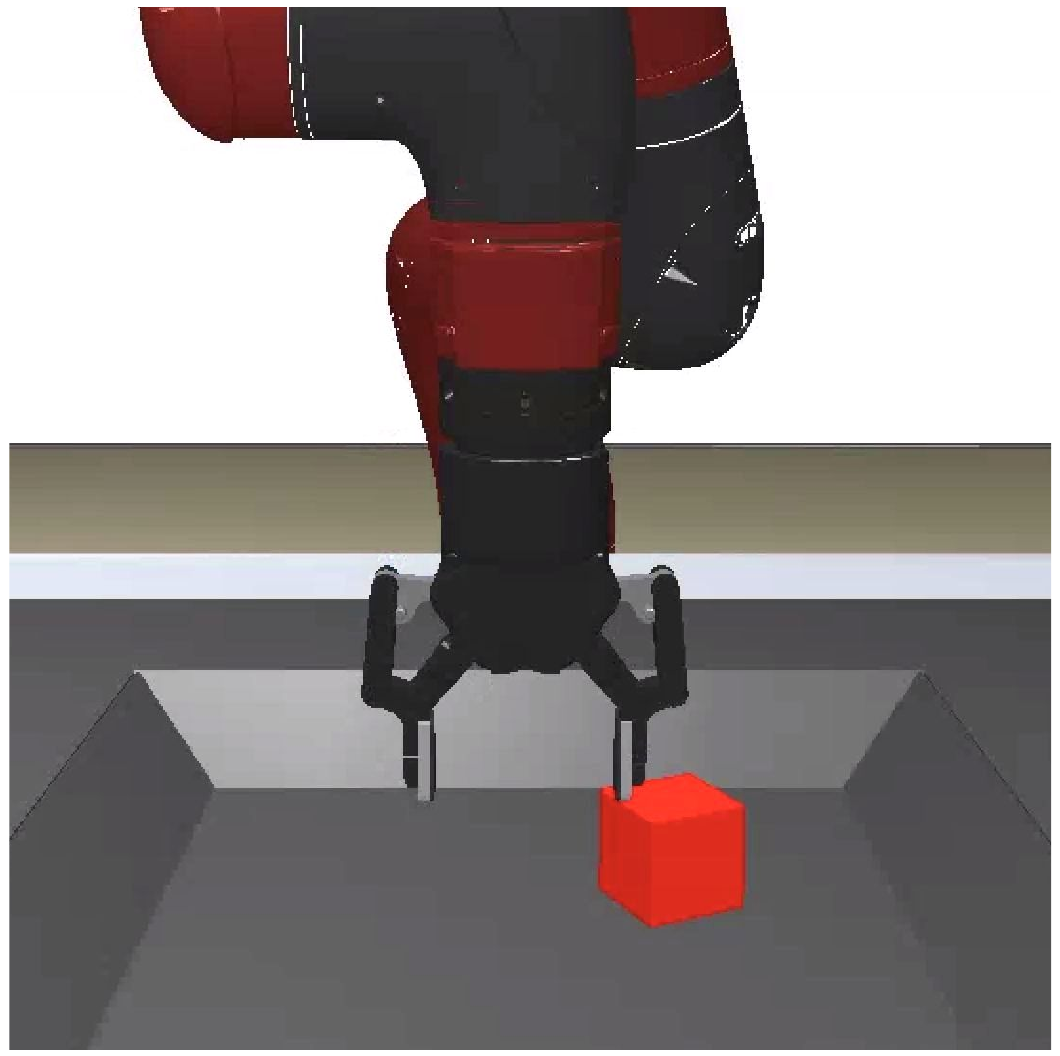}
  \includegraphics[width=0.225\linewidth,height=0.225\linewidth]{3_cube.pdf}
  \caption{Two work spaces, \textit{Jaco} (top) which uses the Jaco arm and is 20 $\times$ 20 cm, and \textit{Sawyer} (bottom) which uses the Sawyer arm and more closely resembles a robot cage and is 35 $\times$ 35 cm.}
  \label{fig:cage_env}
\end{figure}

All our simulations are conducted using MuJoCo\footnote{\scriptsize \url{www.mujoco.org}} \citep{todorov2012mujoco}. We test our proposed algorithms in a variety of different envrionments using simulated Kinova Jaco\footnote{\scriptsize \url{https://www.kinovarobotics.com/en/products/assistive-technologies/kinova-jaco-assistive-robotic-arm}}, and Sawyer robot arms\footnote{\scriptsize \url{https://www.rethinkrobotics.com/sawyer/}} (see Figure~\ref{fig:cage_env}). 
We use the Robotiq 2F85 gripper\footnote{\scriptsize \url{https://robotiq.com/products/2f85-140-adaptive-robot-gripper}} in conjuction with the Sawyer arm. 

To provide demonstrations, we use the SpaceNavigator 3D motion controller\footnote{\scriptsize \url{https://www.3dconnexion.com/spacemouse_compact/en/}} to set Cartesion velocities of the robot arm. The gripper actions are implemented via the buttons on the controller.
All demonstrations in our experiments are provided via human teleoperation and we collected 100 demonstrations for each experiment.

\begin{figure}[ht]
  \centering
  \includegraphics[height=0.225\linewidth]{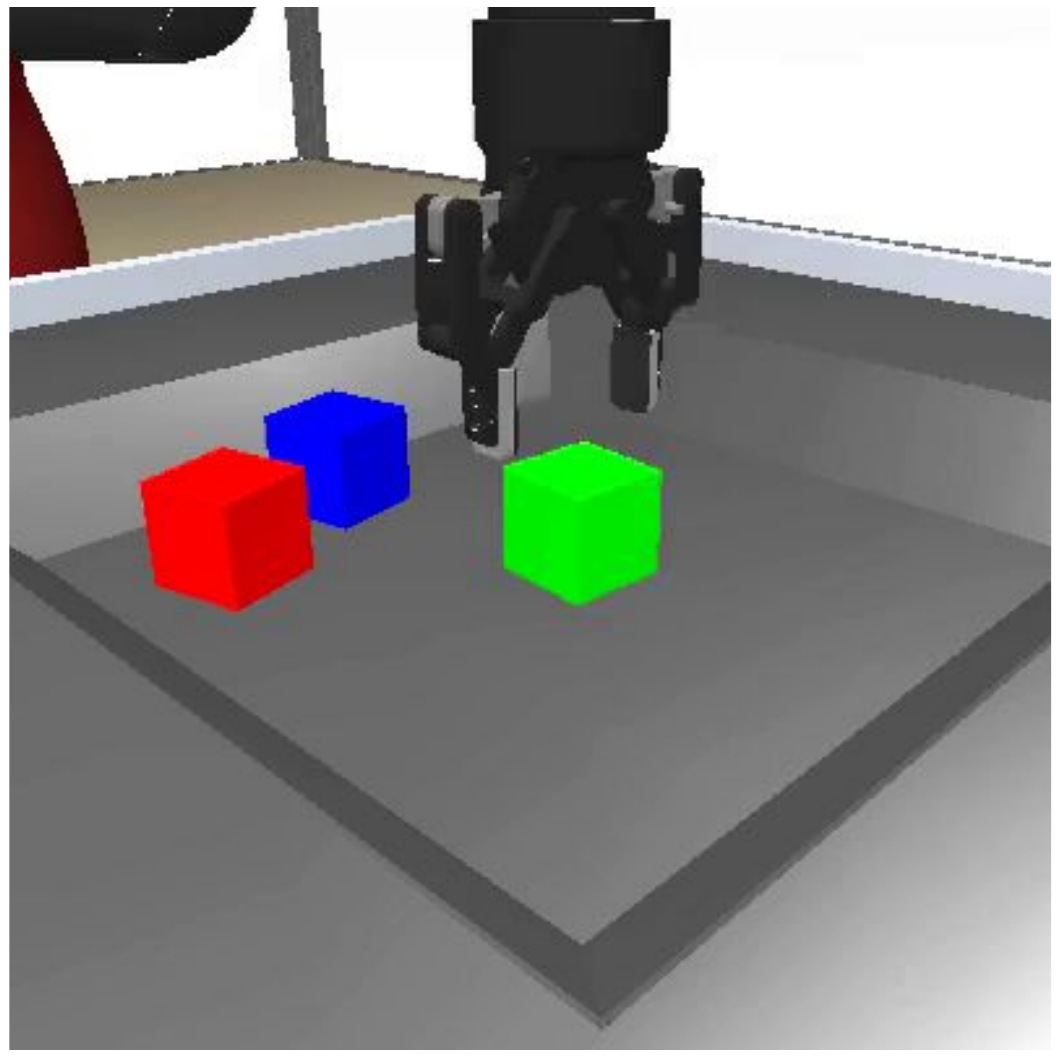}
  \includegraphics[height=0.225\linewidth]{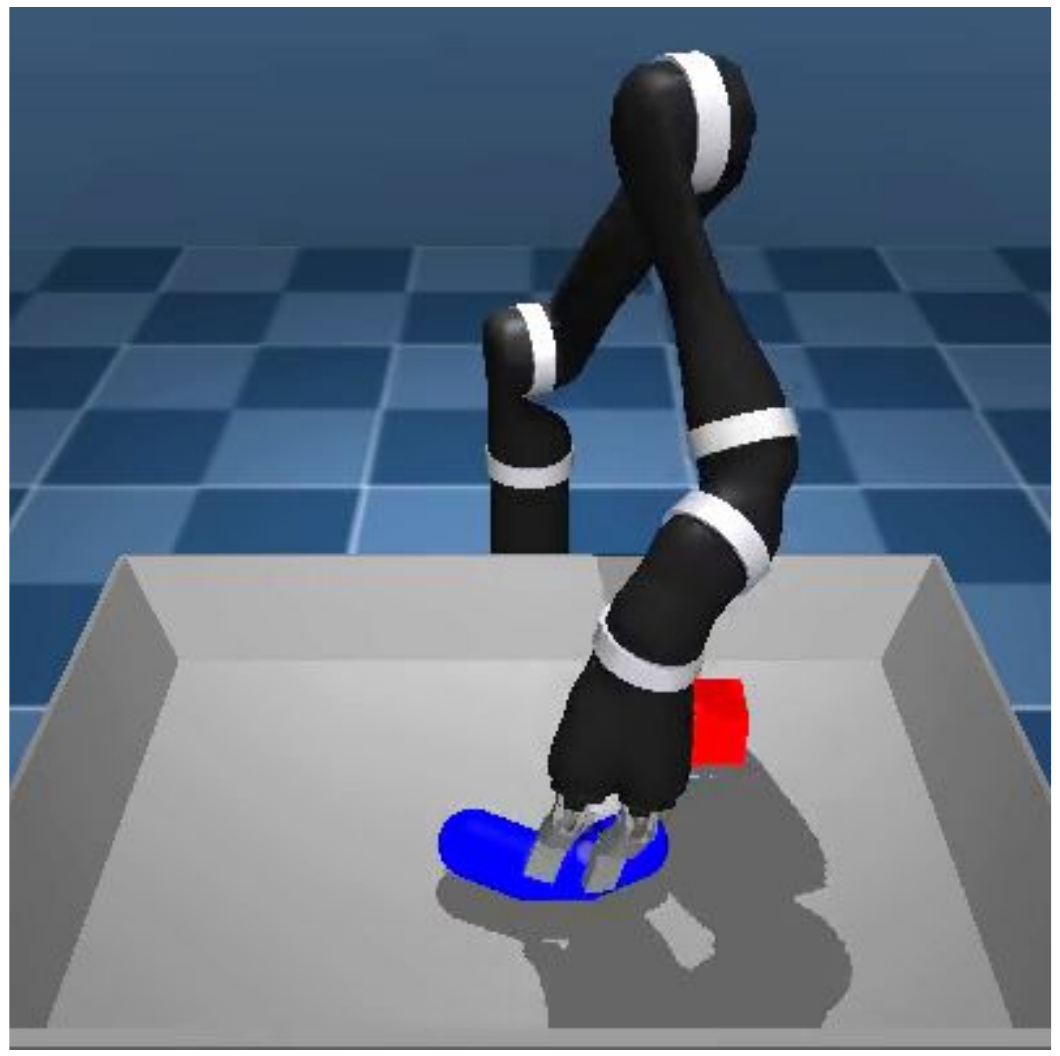}
  \includegraphics[height=0.225\linewidth]{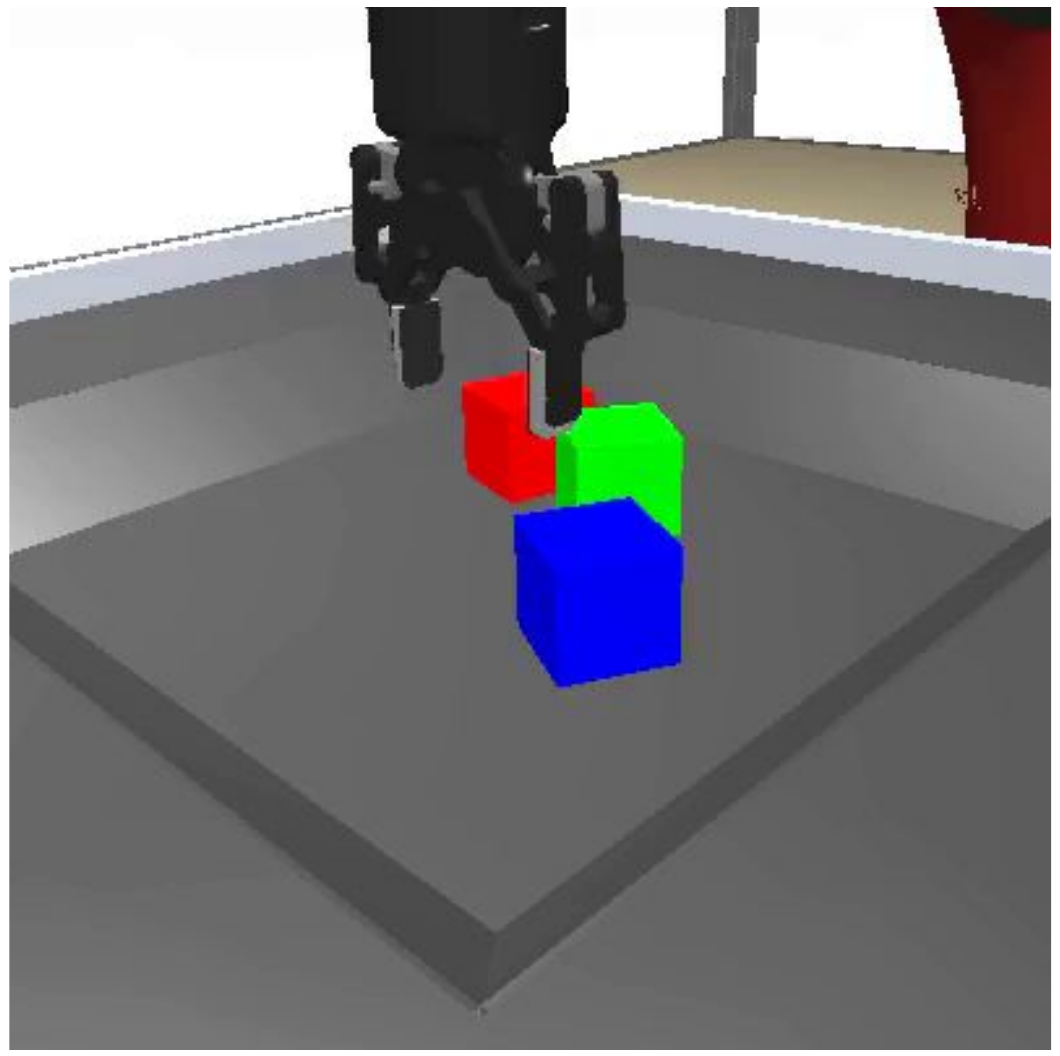}
  \caption{Illustration of the pixels inputs to the agent: front left camera (left), frontal camera (middle) and front right camera (right).}
  \label{fig:cams}
\end{figure}

\textbf{Jaco}: When using the Jaco arm, we use joint velocity control (9DOF) where we control all 6 joints of arm and all 3 joints of the hand.
The simulation is run with a numerical time step of 10 milliseconds, integrating 5 steps, to get a control frequency of 20HZ.
The agent uses a frontal camera of size $64 \times 64$ (see Figure~\ref{fig:cams}).
For a full list of observations the agent sees, please refer to Table~\ref{fig:obs_dims}(a).

\textbf{Sawyer}: When using the Sawyer arm, we use Cartesian velocity control (6DOF) for the robot arm and add one additional action for the gripper resulting in 7 degrees of freedom.
The simulation is run with a numerical time step of 10 milliseconds, integrating 10 steps, to get a control frequency of 10HZ.
The agent uses two frontal cameras of size $64 \times 64$ situated on the left and right side of the robot cage respectively (see Figure~\ref{fig:cams}). For a full list of observations the agent sees, please refer to Table~\ref{fig:obs_dims}(b).

\begin{table}[ht]
  \vspace{-0.0cm}
  \caption{Observation and dimensions.}
  \label{fig:obs_dims}
  \centering
  \footnotesize
  \vspace{0.2cm}
  \footnotesize
  \begin{tabular}{cc}
  \textbf{a) Jaco} & \textbf{b) Sawyer} \\
  \hspace{-1.5em}
  \raisebox{0.58em}{
  \begin{tabular}{l|r}
    \textbf{Feature Name} & \textbf{Dimensions} \\
    \hline \hline
    frontal camera & $64 \times 64 \times 3$ \\
    base force and torque sensors & 6 \\
    arm joints position & 6 \\
    arm joints velocity & 6 \\
    wrist force and torque sensors & 6 \\
    hand finger joints position & 3 \\
    hand finger joints velocity & 3 \\
    hand fingertip sensors & 3 \\
    grip site position &  3 \\
    pinch site position & 3 \\
    \hline \hline
  \end{tabular}
  } &
  \begin{tabular}{l|r}
    \textbf{Feature Name} & \textbf{Dimensions} \\
    \hline \hline
    front left camera & $64 \times 64 \times 3$ \\
    front right camera & $64 \times 64 \times 3$ \\
    arm joint position & 7 \\
    arm joint velocity & 7\\
    wrist force sensor & 3\\
    wrist torque sensor & 3\\
    hand grasp sensor & 1\\
    hand joint position & 1\\
    tool center point cartesian orientation & 9\\
    tool center point cartesian position & 3\\
    hand joint velocity & 1 \\
    \hline \hline
  \end{tabular}
  \end{tabular}
\end{table}

For all environments considered in this paper, we provide sparse rewards (\emph{i.e.}. if task is accomplished, the reward is 1 and 0 otherwise). 
In experiments regards our proposed methods, rewards are only used for evaluation purposes and not for training the agent.

\section{Data augmentation}\label{sec:data_augmentation}

Traditional data augmentation has proved beneficial in imitation learning \citep{berseth2019visual}.
Surprisingly, this has received little attention in prior publications on GAIL.
However, we find that data augmentation is a necessary procedure to prevent discriminator overfitting.
Therefore, all discriminator-based methods in this work (including all baselines) use data augmentation with an exception of the experiments in this section.
We distort images by randomly changing brightness, contrast and saturation, randomly cropping and rotating, and adding Gaussian noise.
When multiple sensor inputs are available (e.g. multiple cameras), we also randomly drop some of the inputs, always leaving at least one active.

We also considered regularizing the GAIL discriminator with spectral normalization~\citep{miyato2018spectral}.
It performed slightly better than GAIL, but still failed in the presence of distractor objects, and we thus omit spectral normalization in the main experiments for simplicity.

\begin{figure}[th!]
  \centering
  \includegraphics[height=0.225\linewidth]{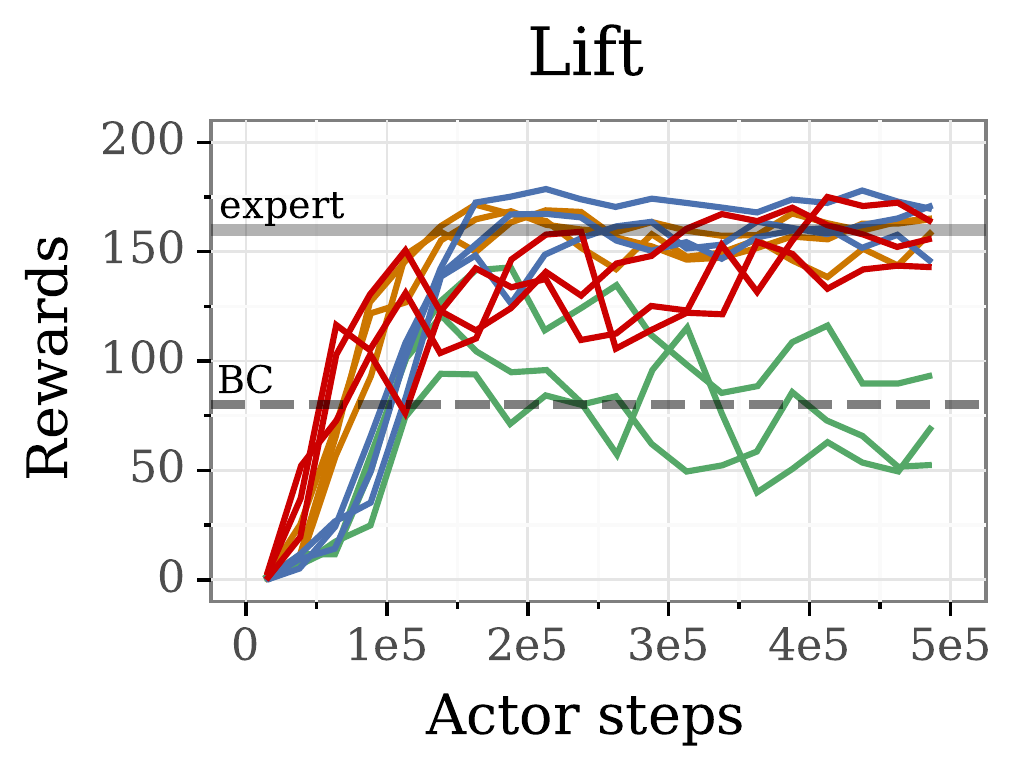} \hspace{-0.15cm}
  \includegraphics[height=0.225\linewidth]{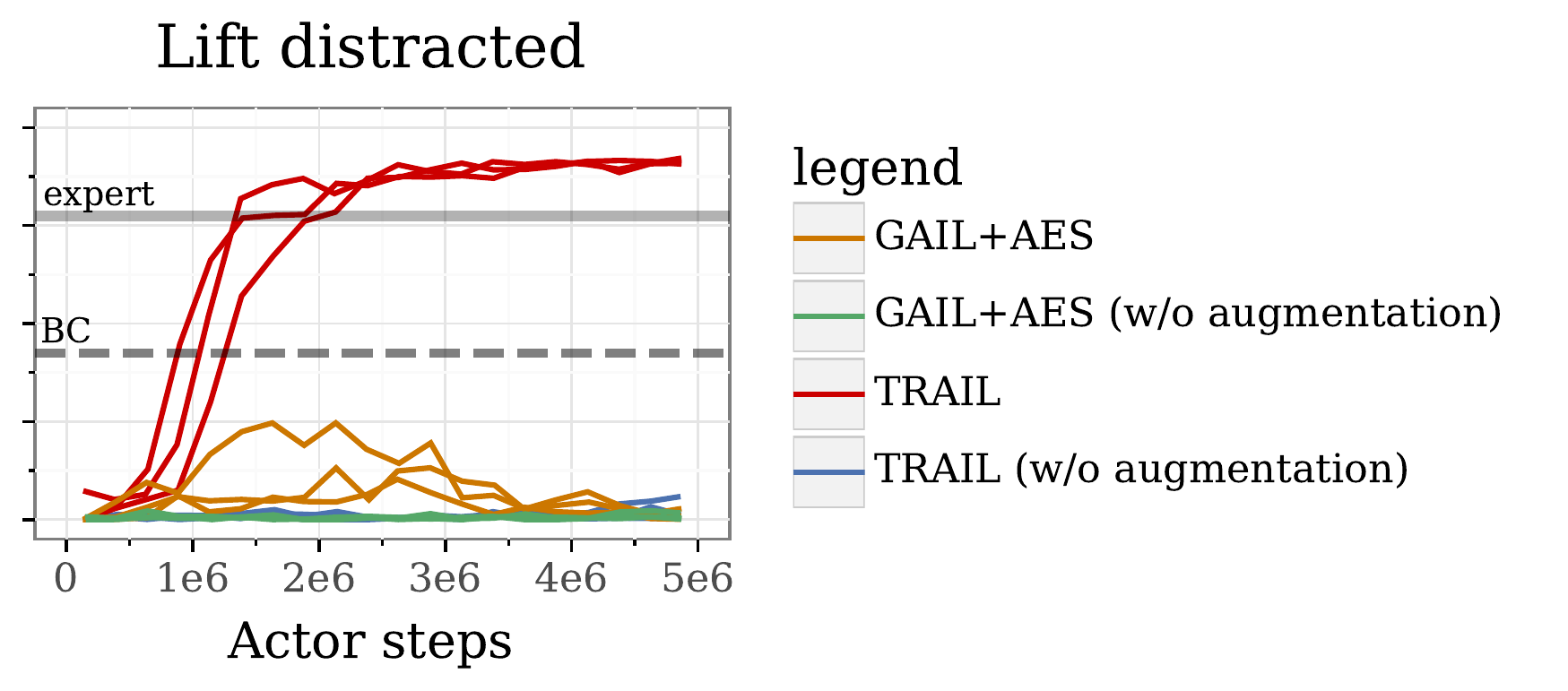}
  \caption{Results for \textit{lift} and \textit{lift distracted} in \textit{Sawyer} work space. TRAIL and \aes{} are by default with data augmentation. Additional results without data augmentation are presented to show its importance.}
  \label{fig:results_augmentation}
\end{figure}

\begin{wraptable}{r}{0.55\textwidth}
  \vspace{-0.15in}
  \caption{Influence of data augmentation (evaluated on rewards) for \textit{lift} and \textit{lift distracted}.} 
  \footnotesize
  \label{tab:data-augmentation}
  \centering
  \begin{tabular}{c|c|cc}
    \toprule
    \multirow{2}{*}{Task} & \multirow{2}{*}{Regularization}  & \multicolumn{2}{c}{Data augmentation} \\
    & & Yes & No \\
    \midrule
    \multirow{2}{*}{\textit{lift}} & \aes{} & $\sim$165 & $\sim$115 \\
     & TRAIL & $\sim$155 & $\sim$165 \\
    \midrule
    \multirow{2}{*}{\textit{lift distracted}} & \aes{} & $\sim$30 & $\sim$5 \\
     & TRAIL & $\sim$180 & $\sim$10 \\
    \bottomrule
  \end{tabular}
  \vspace{-0.15in}
\end{wraptable}
In this section, we measure the importance of data augmentation. In Table~\ref{tab:data-augmentation}, we report the best reward obtained in the first 12 hours of training (averaged for all seeds; see Figure~\ref{fig:results_augmentation} for full curves).

The results show that data augmentation is essential in \textit{lift distracted}. The performance of TRAIL is not affected by the lack of data augmentation on the \emph{lift} task, whereas the performance of \aes{} is.

We also checked the impact of data augmentation on the baseline GAIL. It was  always beneficial. Since the baseline GAIL which uses data augmentation usually does not take off, the even worse version is not included in our plots.

\section{Block lifting with distractors}\label{sec:lift_distracted_abl}

\begin{figure}[h]
  \centering
  \includegraphics[height=0.225\linewidth]{lift_alone} \hspace{-0.15cm}
  \includegraphics[height=0.225\linewidth]{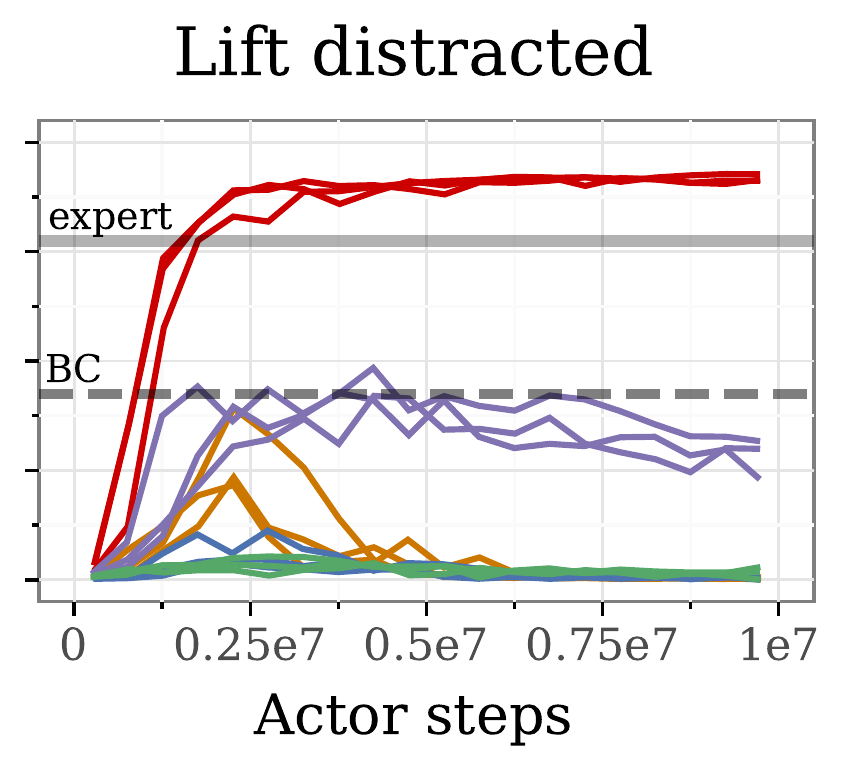} \hspace{-0.15cm}
  \includegraphics[height=0.225\linewidth]{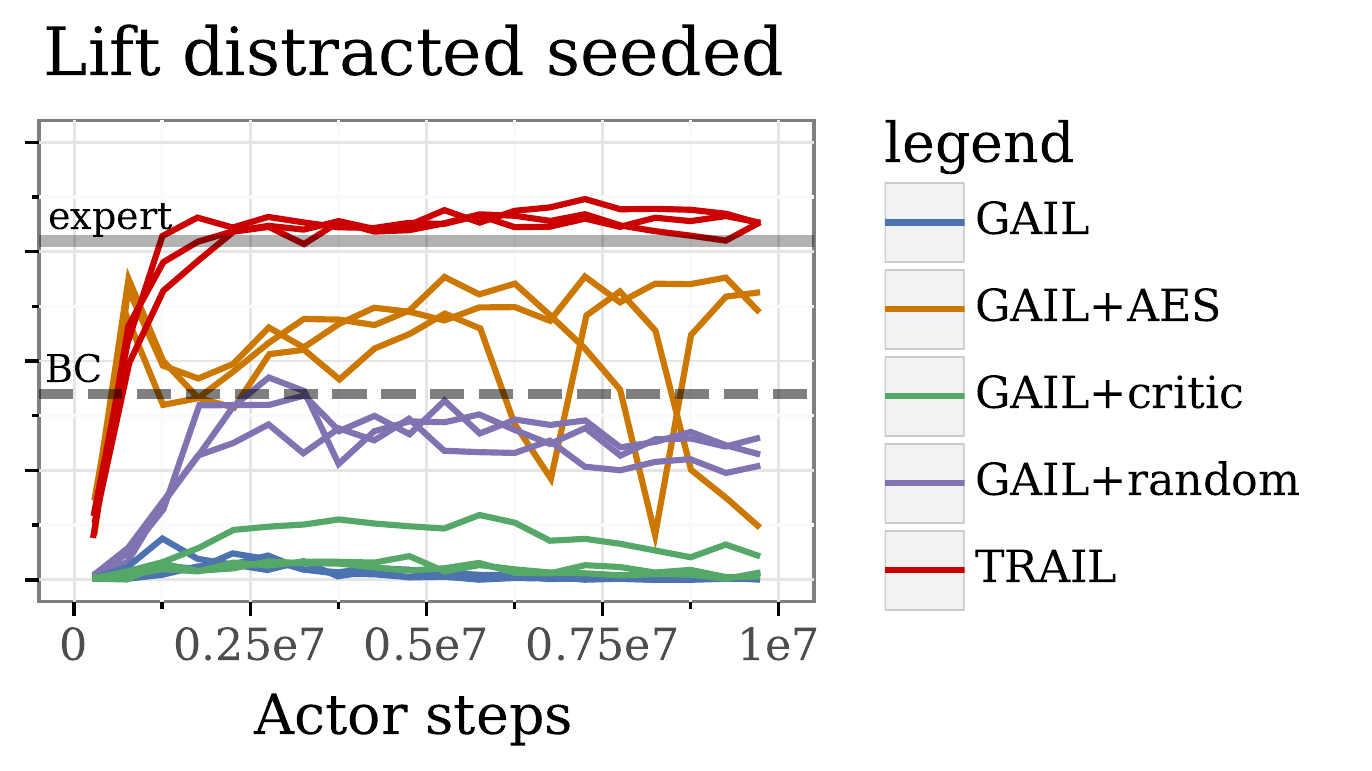}
  \caption{Results for \textit{lift}, \textit{lift distracted}, and \textit{lift distracted seeded}. Only TRAIL excels.}
  \label{fig:rgb_lift_full}
\end{figure}

In Section~\ref{sec:baselines}, we consider a variant of the lift task, \textit{lift distracted}, when two extra blocks are added.
We also conducted an additional experiment termed \textit{lift distracted seeded}, in which the initial block positions are randomly drawn from the expert demonstrations.
Therefore, it prevents from discriminating between expert and actor episodes using distractor positions.
Note this initialization procedure is not applied at the evaluation time, keeping the evaluation scores comparable between \textit{lift distracted} and \textit{lift distracted seeded}.
The results for all lift variants are presented in Figure~\ref{fig:rgb_lift_full}.

All methods perform satisfactorily on \textit{lift}, but the proposed methods, \aes{} and TRAIL, clearly outperform the other methods.

As expected, the performance of BC on \textit{lift distracted} is similar to its performance on \textit{lift}, despite the two additional blocks.
The two additional blocks, however, affect all GAIL-based baselines, despite being irrelevant to the task. Only TRAIL succeeds.

It turns our that even for this relatively easy task in the simulated environment without evident task-irrelevant features, the baseline discriminators find spurious associations to obtain perfect accuracy.
It is achieved by simply memorizing all 100 initial block positions from the demonstration set.
Note that this can not be prevented using unstructured regularization, as a discriminator has to be able to infer blocks' positions from the image to provide informative reward.
However, the positions of task-irrelevant objects should be ignored.
TRAIL is the only method that is able to handle the variety of initial cube positions, achieving better than expert performance on \textit{lift distracted}.

\aes{} performed much better on \textit{lift distracted seeded} than on \textit{lift distracted}, indicating that the discriminator do form a spurious association between distractor positions and expert labeling.

\begin{figure}[th!]
  \centering
  \includegraphics[height=0.225\linewidth]{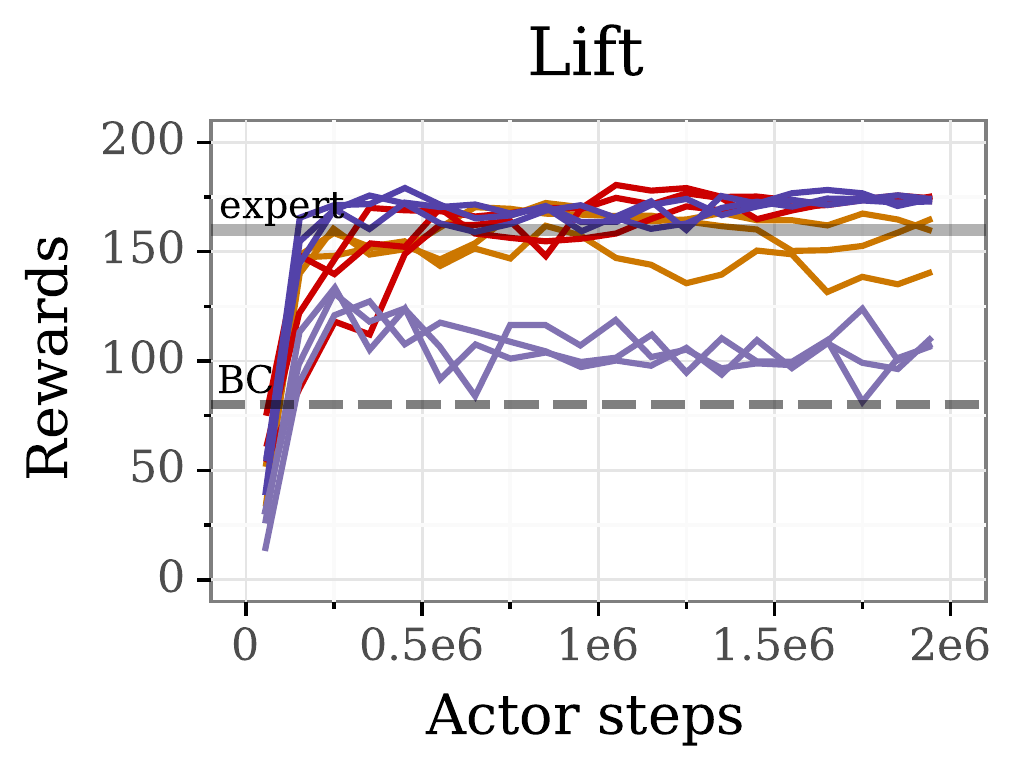} \hspace{-0.15cm}
  \includegraphics[height=0.225\linewidth]{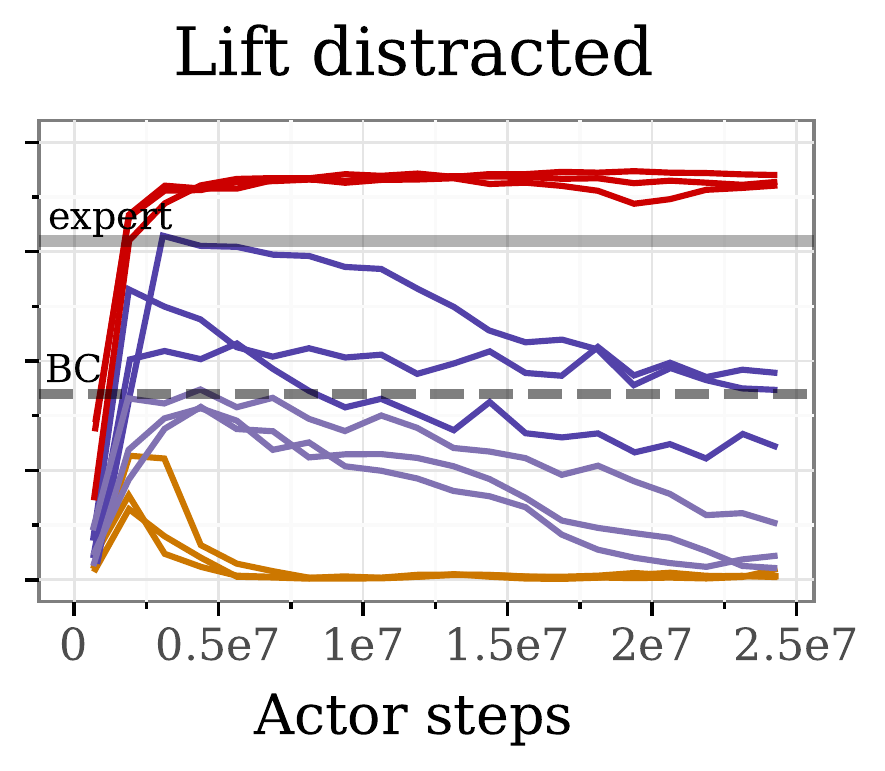} \hspace{-0.15cm}
  \includegraphics[height=0.225\linewidth]{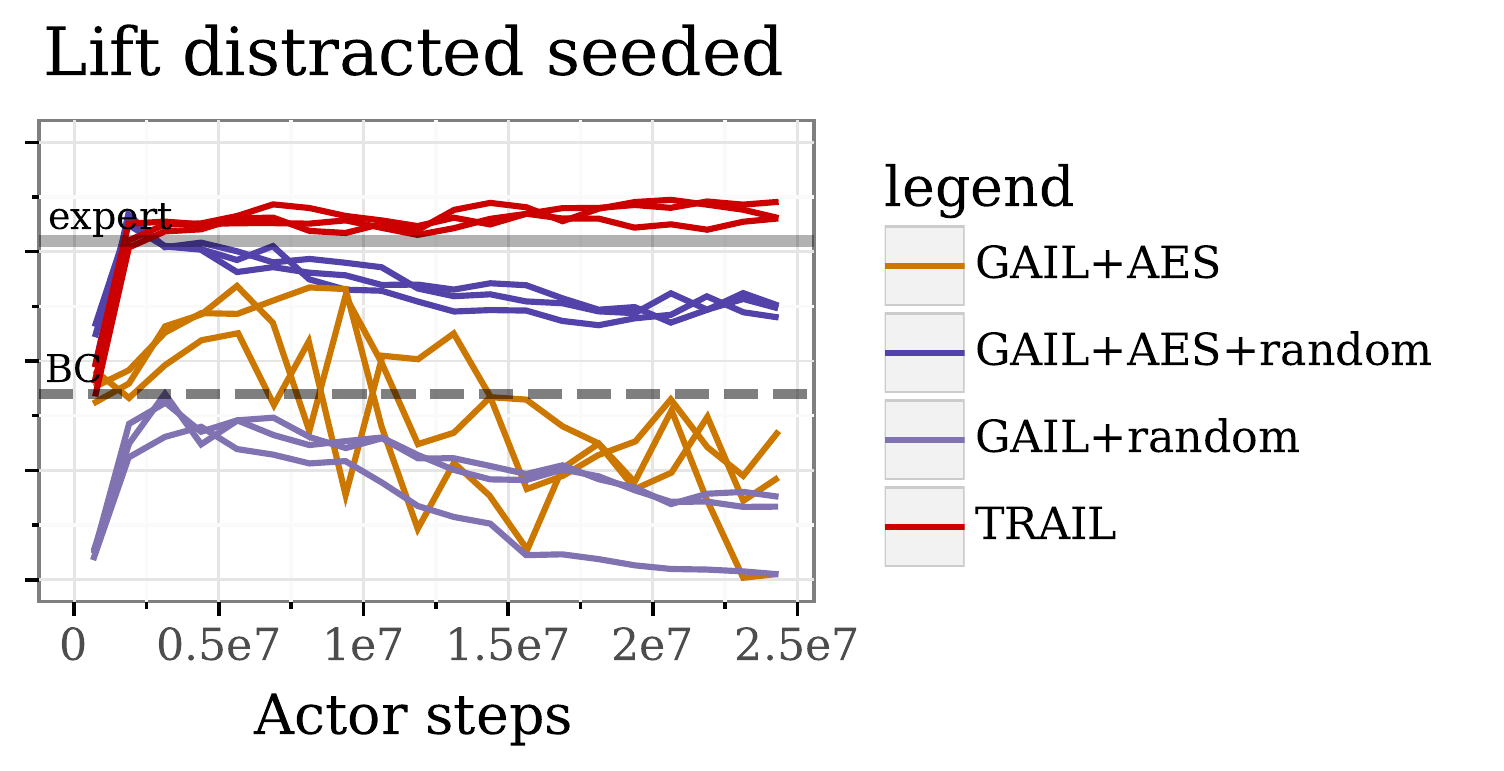}
  \caption{Results for \textit{lift}, \textit{lift distracted}, and \textit{lift distracted seeded}
  \label{fig:rgb_lift_red_r} including \mbox{~\aes{}+random} baseline.}
\end{figure}

Interestingly, GAIL+random performs reasonably on both \textit{lift distracted} and \textit{lift distracted seeded}.
Given GAIL+random's strong performance, we conducted additional experiments to evaluate its effectiveness when trained with adaptive early stopping.
In Figure~\ref{fig:rgb_lift_red_r} we present \mbox{~\aes{}+random}, which is GAIL-random trained with actor early stopping.

When random features are used it is harder to identify and exploit the spurious associations.
Indeed, \mbox{~\aes{}+random} and TRAIL are the only methods exceeding BC performance on \textit{lift distracted}.
However, TRAIL performance is clearly better (obtains higher rewards and never gets overfitted) which means that \mbox{~\aes{}+random} is still prone to exploiting spurious associations.

\section{Early frames ablation study}\label{sec:early_frames}

\begin{figure}[th!]
  \centering
  \vspace{-0.1in}
  \includegraphics[height=0.225\linewidth]{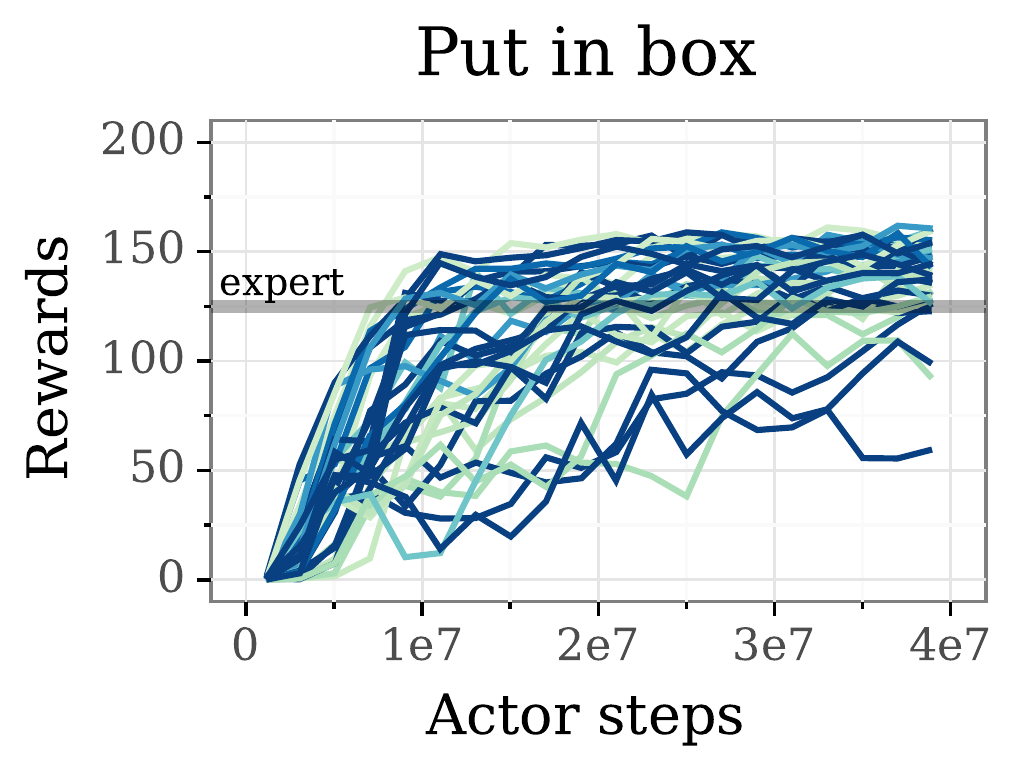}
  \hspace{-0.15cm}
  \includegraphics[height=0.225\linewidth]{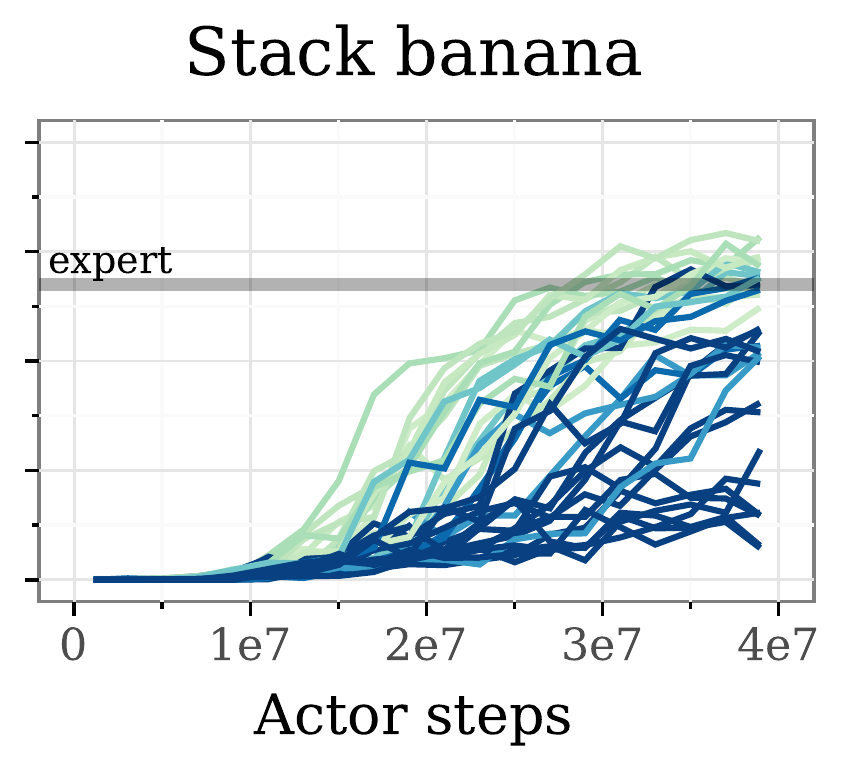}
  \hspace{-0.15cm}
  \includegraphics[height=0.225\linewidth]{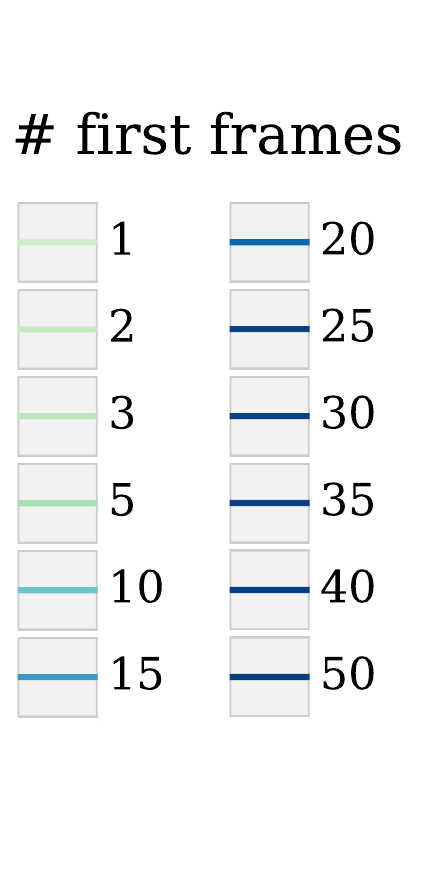}
  \caption{TRAIL performance, varying the number of first frames in each episode used to form $\mathcal{I}$.
  \label{fig:ablation_size}}
\end{figure}

Here, we vary the number of early frames of each episode used to form the constraining sets. We report that a range of values from 1 up to 20 works well across both tasks (Put in box and Stack banana), with performance gradually degrading as the value increased beyond 20 (Figure~\ref{fig:ablation_size}).

\section{Fixed termination policy}\label{sec:fixed_termination}

\begin{figure}[h]
  \centering
  \vspace{-0.1in}
  \includegraphics[height=0.225\linewidth]{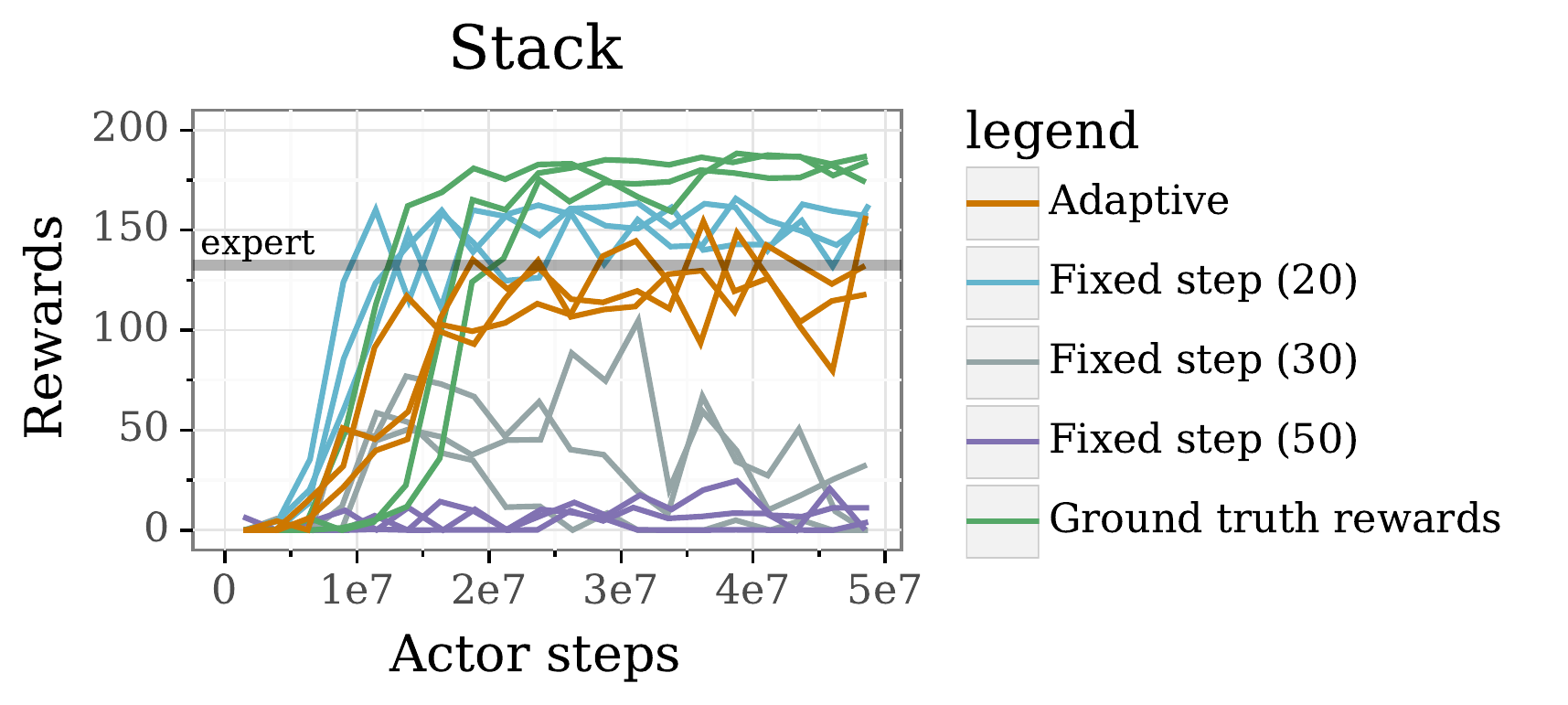}
  \caption{Results for \textit{stack} in \textit{Jaco} work space. Fixed step termination policy can be very effective but the final performance is very sensitive to the hyperparameter. \aes{} does not need tuning nor access to the environment reward.}
  \label{fig:fixed_step}
\end{figure}

As mentioned in the subsection~\ref{sec:experiments_early_stopping}, the most basic early termination policy -- fixed step termination -- may be very effective if tuned. Since the tuning may be expensive in practice, we recommend using adaptive early stopping (\aes). However, for the sake of completeness we provide results for fixed step termination policy depending on the hyperparameter tuned. The results for \textit{stack} task are presented in Figure~\ref{fig:fixed_step}. As can be inferred from the figure, the performance is very sensitive to the fixed step hyperparameter. We refer to Section~\ref{sec:experiments_early_stopping} for more comments on all methods.

\section{Learning with oracle discriminator}
\label{sec:oracle_discriminator}

\begin{figure}[h]
  \centering
  \vspace{-0.1in}
  \includegraphics[height=0.225\linewidth]{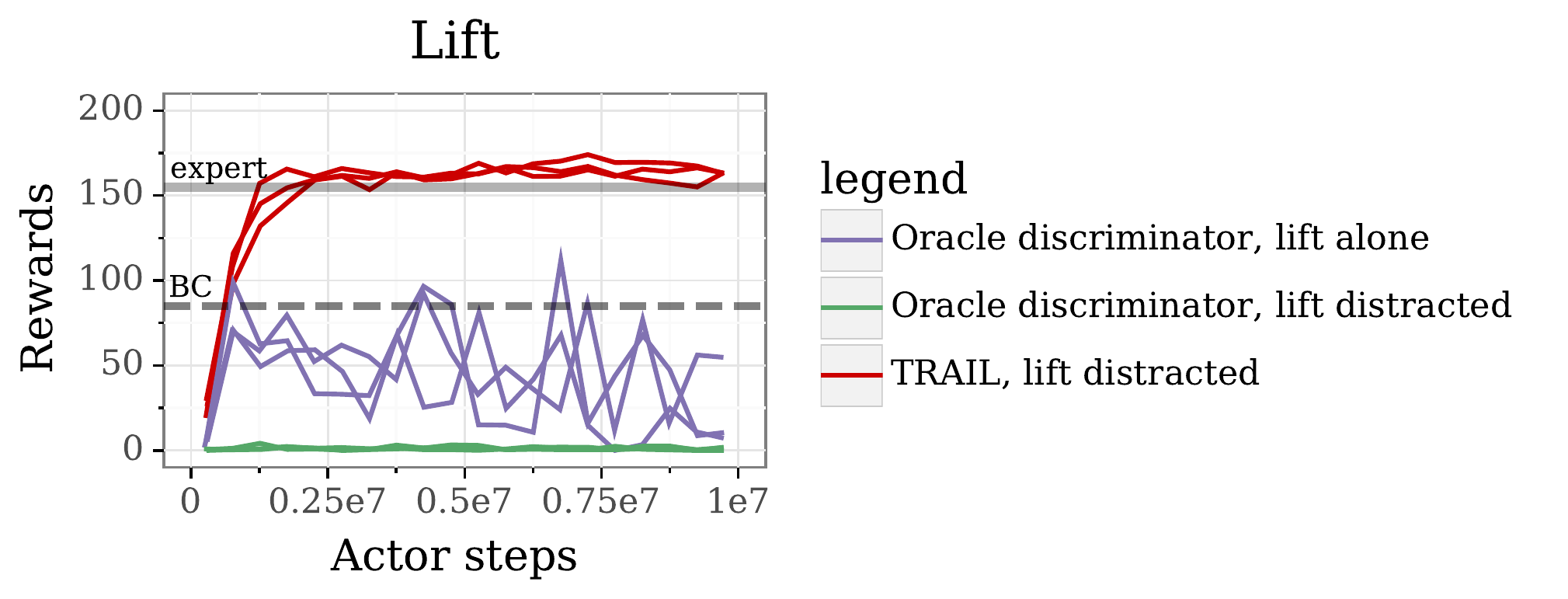}
  \caption{With fixed rewards, the agent is able to learn \textit{lift} somewhat, but performs worse than TRAIL. When distractor blocks are added, the fixed reward agent fails to learn completely. \label{fig:fixed_discriminator}}
  \vspace{-0.1in}
\end{figure}

To further understand the contribution of learned reward functions, we train agents with rewards from an oracle discriminator, which always assigns a reward of $1$ for expert frames and $0$ for agent frames.
This simulates a discriminator that perfectly exploits a spurious association between task-irrelevant visual features and expert labels.
On the \textit{lift} task, agents using this fixed oracle reward achieve roughly half the reward of TRAIL asymptotically, and on \textit{lift distracted} they do not solve the task (average rewards are less than 5).
The learning curves are presented in Figure~\ref{fig:fixed_discriminator}.

\section{D4PG}
\label{appendix:d4pg}
We use D4PG~\citep{barth2018distributed} as our main training algorithm. Briefly, D4PG is a distributed off-policy reinforcement learning algorithm for continuous control problems.
In a nutshell, D4PG uses Q-learning for policy evaluation and Deterministic Policy Gradients (DPG) \citep{silver2014deterministic} for policy optimization.
An important characteristic of D4PG is that it maintains a replay memory $\cal{M}$ (possibility prioritized \citep{horgan2018distributed}) that stores SARS tuples which allows for off-policy learning.
D4PG also adopts target networks for increased training stability.
In addition to these principles, D4PG utilized distributed training, distributional value functions, and multi-step returns to further increase efficiency and stability. In this section, we explain the different ingredients of D4PG.

D4PG maintains an online value network $Q(s, a | \theta)$ and an online policy network $\pi(s | \phi)$. The target networks are of the same structures as the value and policy network, but are parameterized by different parameters $\theta'$ and $\phi'$ which are periodically updated to the current parameters of the online networks.

Given the $Q$ function, we can update the policy using DPG:
\begin{align}
\mathcal{J}(\phi) = \mathbb{E}_{s_t \sim \cal{M}}\big[\nabla_{\phi} Q(s_t, \pi(s_t | \phi) | \theta) \big]. \label{eqn:dpg}
\end{align}
Instead of using a scalar $Q$ function, D4PG adopts a distributional value function such that
$Q(s_t, a|\theta) = \mathbb{E} \big[ Z(s_t, a|\theta) \big]$ where $Z$ is a random variable such that $Z = z_i$ w.p. $p_i \propto \exp(\omega(s_t, a|\theta))$. The $z_i$'s take on $V_{bins}$ discrete values that ranges uniformly between $V_{min}$ and $V_{max}$ such that $z_i = V_{min} + i\frac{V_{max} - V_{min}}{V_{bins}}$ for $i \in \{0, \cdots, V_{bins}-1\}$.

To construct a bootstrap target, D4PG uses N-step returns. Given a sampled tuple from the replay memory: $s_t, a_t, \{r_t, r_{t+1}, \cdots, r_{t+N-1}\}, s_{t+N}$, we construct a new random variable $Z'$ such that 
$Z' = z_i + \sum_{n=0}^{N-1} \gamma^n r_{t+n}$ w.p. $p_i \propto \exp(\omega(s_{t+N}, \pi(s_{t+N}|\phi') | \theta'))$. 
Notice, $Z'$ no longer has the same support. We therefore adopt the same projection $\Phi$ employed by \cite{bellemare2017distributional}.
The training loss for the value function
\begin{align}
\mathcal{L}(\theta) = \mathbb{E}_{s_t, a_t, \{r_t, \cdots, r_{t+N-1}\}, s_{t+N} \sim \cal{M}} \big[ H\big(\Phi(Z'), Z(s_t, a_t | \theta)\big) \big], \label{eqn:dist_q}
\end{align}
where $H$ is the cross entropy.

D4PG is also distributed following \cite{horgan2018distributed}. Since all learning processes only rely on the replay memory, we can easily decouple the `actors' from the `learners'. D4PG therefore uses a large number of independent actor processes which act in the environment and write data to a central replay memory process. The learners could then draw samples from the replay memory for learning. The learner also serves as a parameter server to the actors which periodically update their policy parameters from the learner.

In our experiments, we always have access to expert demonstrations. We, therefore adopt the practice from DQfD and DDPGfD and put the demonstrations into our replay buffers.
For more details see Algorithms 
\ref{alg:actor} and \ref{alg:learner}.

\begin{algorithm}[H]
\caption{Actor\label{alg:actor}}
\begin{algorithmic} 
\STATE \textbf{Given:}
an experience replay memory $\mathcal{M}$
\FOR {$n_{episodes}$}
\FOR {$t=1$  {\bfseries to} $T$}
\STATE Sample action from task policy: $a_t \leftarrow \pi(s_t)$
\STATE Execute action $a_t$ and observe new state $s_{t+1}$, and reward $r_t$.
\STATE Store transition $(s_t, a_t, r_t, s_{t+1}$) in memory $\mathcal{M}$
\ENDFOR
\ENDFOR
\end{algorithmic}
\end{algorithm}

\begin{algorithm}[H]
\caption{Learner\label{alg:learner}}
\begin{algorithmic} 
\STATE \textbf{Given:} an off-policy RL algorithm $\mathbb{A}$, a replay buffer $\cal{M}$, a replay buffer of expert demonstrations ${\cal{M}}_e$
\STATE Initialize $\mathbb{A}$
\FOR {$n_{updates}$}
  \STATE Sample transitions $(s_t, a_t, r_t, s_{t+1})$
  from $\mathcal{M}$ to make a minibatch $B$.
  \STATE Sample transitions $(s_t, a_t, r_t, s_{t+1})$
  from ${\cal{M}}_e$ enlarge the minibatch $B$.
  \STATE Perform a actor update step with Eqn. (\ref{eqn:dpg}).
  \STATE Perform a critic update step with Eqn. (\ref{eqn:dist_q}).
  \STATE Update the target actor/critic networks every $k$ steps.
\ENDFOR
\end{algorithmic}
\end{algorithm}

\section{Network architecture and hyperparameters}
\label{sec:hyperparameters}

Actor and critic share a residual pixel encoder network with eight convolutional layers (3x3 convolutions, three 2-layer blocks with 16, 32, 32 channels), instance normalization \citep{ulyanov2016instance} and exponential linear units \citep{clevert2015fast} between layers.

\begin{figure}[h!]
  \centering
  \includegraphics[width=\linewidth]{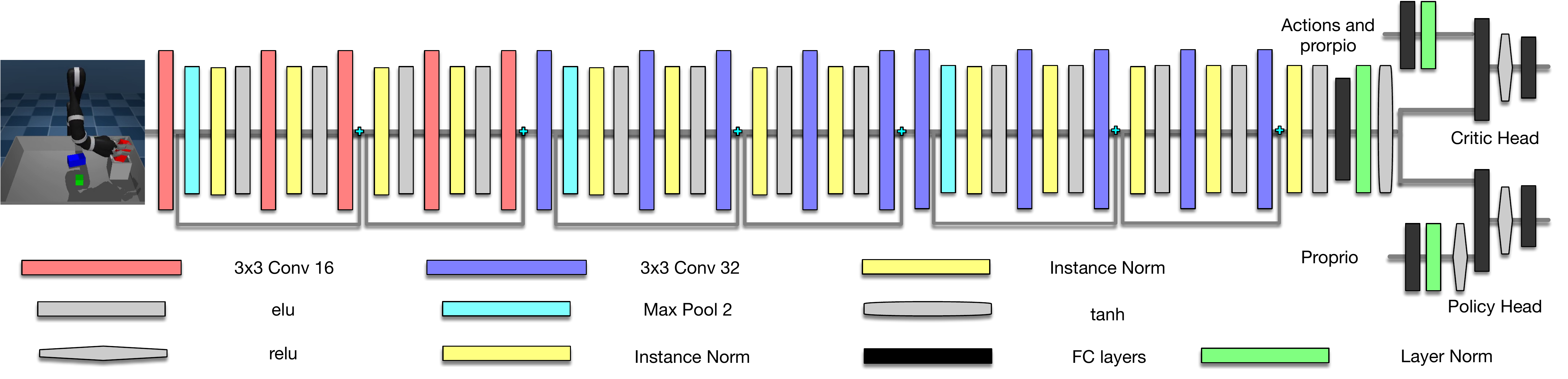}
  \vspace{-0.3cm}
  \caption{Network architecture for the policy and critic. \label{fig:network_trail}}
  \vspace{-0.1cm}
\end{figure}

The policy is a 3-layer MLP with ReLU activations with hidden layer sizes (300, 200).
The critic is a 3-layer MLP with ReLU activations with hidden layer sizes (400, 300).
For a illustration of the network. Please see \autoref{fig:network_trail}.
The discriminator network uses a pixel encoder of the same architecture as the actor critic, followed by a 3-layer MLP with ReLU activations and hidden layer sizes (32, 32). Table~\ref{tab:hyperparams} shows all network hyper parameters.

\begin{table}[!h]
\centering
\caption{Hyper parameters used in all experiments.\label{tab:hyperparams}}
\begin{tabular}{@{}lll@{}}\toprule
\textbf{Parameters} & \textbf{Values}\\

\midrule
Actor/Critic Input Width and Height & $64 \times 64$\\

\midrule
\textbf{Actor-Critic Parameters} &\\
\hphantom{---}$V_{min}$ & $-50$\\
\hphantom{---}$V_{max}$ & $150$\\
\hphantom{---}$V_{bins}$ & $21$\\
\hphantom{---}N step & $1$\\
\hphantom{---}Learning rate & $10^{-4}$\\
\hphantom{---}Optimizer & Adam~\cite{kingma2014adam}\\
\hphantom{---}Batch size & $256$\\
\hphantom{---}Target update period & $100$\\
\hphantom{---}Discount factor ($\gamma$)& $0.99$\\
\hphantom{---}Replay capacity & $10^6$ &\\
\hphantom{---}Number of actors & $32$ or $128$ &\\

\midrule
\textbf{Imitation Parameters} &\\
\hphantom{---}Discriminator learning rate & $10^{-4}$\\
\hphantom{---}Discriminator Input Width & $48$\\
\hphantom{---}Discriminator Input Height & $48$\\
\bottomrule
\end{tabular}
\end{table}

\end{bibunit}

\end{document}